\documentclass{svjour3}
\usepackage{multirow}
\usepackage[utf8]{inputenc}
\usepackage[T1]{fontenc}
\usepackage{lmodern} 
\usepackage[bitstream-charter]{mathdesign}
\usepackage{inconsolata}
\usepackage{listings}
\usepackage{amsmath}
\usepackage{graphicx}
\usepackage{longtable}
\usepackage{algorithm2e}
\usepackage{hyperref}
\usepackage{booktabs}
\usepackage[frozencache=true]{minted}

\newcommand{\tw}[1]{\texttt{#1}}

\usepackage{xcolor}

\journalname{Machine learning}
\title{Inductive general game playing}
\author{Andrew Cropper \and Richard Evans\thanks{We are very grateful to the following for feedback and guidance throughout this project: Alessandra Russo, David Pfau, Edward Grefenstette, Krysia Broda, Marc Lanctot, Marek Sergot, and Pushmeet Kohli.} \and Mark Law}
\begin{document}

\institute{A. Cropper\at
              University of Oxford, UK\\
              \email{andrew.cropper@cs.ox.ac.uk}
            \and
            R. Evans\at
            Imperial College London, UK\\
              \email{richardevans@google.com}
            \and
                M. Law \at
                Imperial College London, UK\\
              \email{mark.law09@imperial.ac.uk}
}


\maketitle

\begin{abstract}
General game playing (GGP) is a framework for evaluating an agent's general intelligence across a wide range of tasks.
In the GGP competition, an agent is given the rules of a game (described as a logic program) that it has never seen before.
The task is for the agent to play the game, thus generating game traces.
The winner of the GGP competition is the agent that gets the best total score over all the games.
In this paper, we invert this task: a learner is given game traces and the task is to learn the rules that could produce the traces.
This problem is central to \emph{inductive general game playing} (IGGP).
We introduce a technique that automatically generates IGGP tasks from GGP games.
We introduce an IGGP dataset which contains traces from 50 diverse games, such as \emph{Sudoku}, \emph{Sokoban}, and \emph{Checkers}.
We claim that IGGP is difficult for existing inductive logic programming (ILP) approaches.
To support this claim, we evaluate existing ILP systems on our dataset.
Our empirical results show that most of the games cannot be correctly learned by existing systems.
The best performing system solves only 40\% of the tasks perfectly.
Our results suggest that IGGP poses many challenges to existing approaches.
Furthermore, because we can automatically generate IGGP tasks from GGP games, our dataset will continue to grow with the GGP competition, as new games are added every year.
We therefore think that the IGGP problem and dataset will be valuable for motivating and evaluating future research.
\end{abstract}

\section{Introduction}
\label{sec:intro}

\begin{figure}[ht]
\centering
\begin{tabular}{ccc}
\includegraphics[scale=0.12,trim=0 100 0 100, clip]{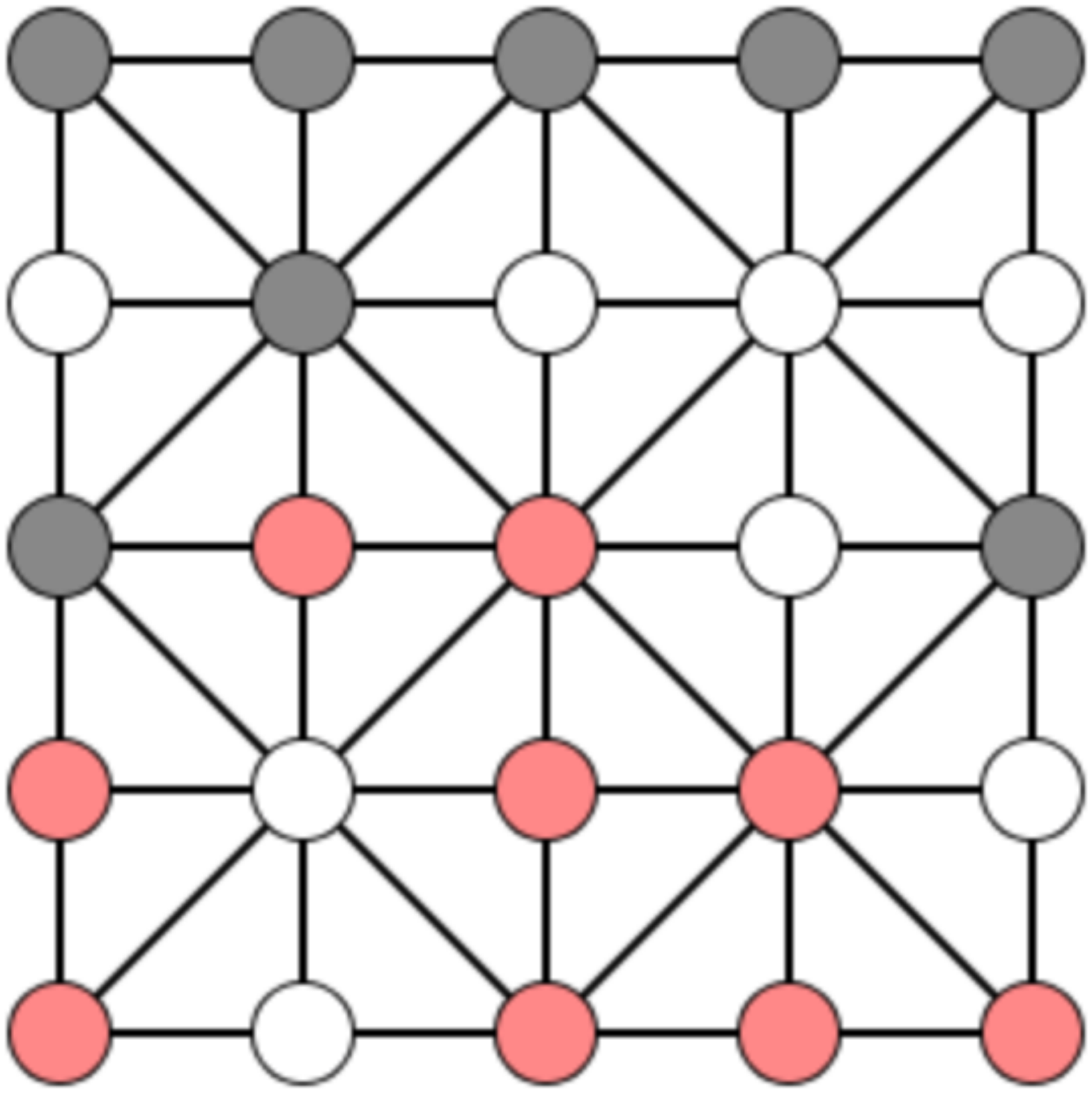} &
\includegraphics[scale=0.12,trim=0 100 0 100, clip]{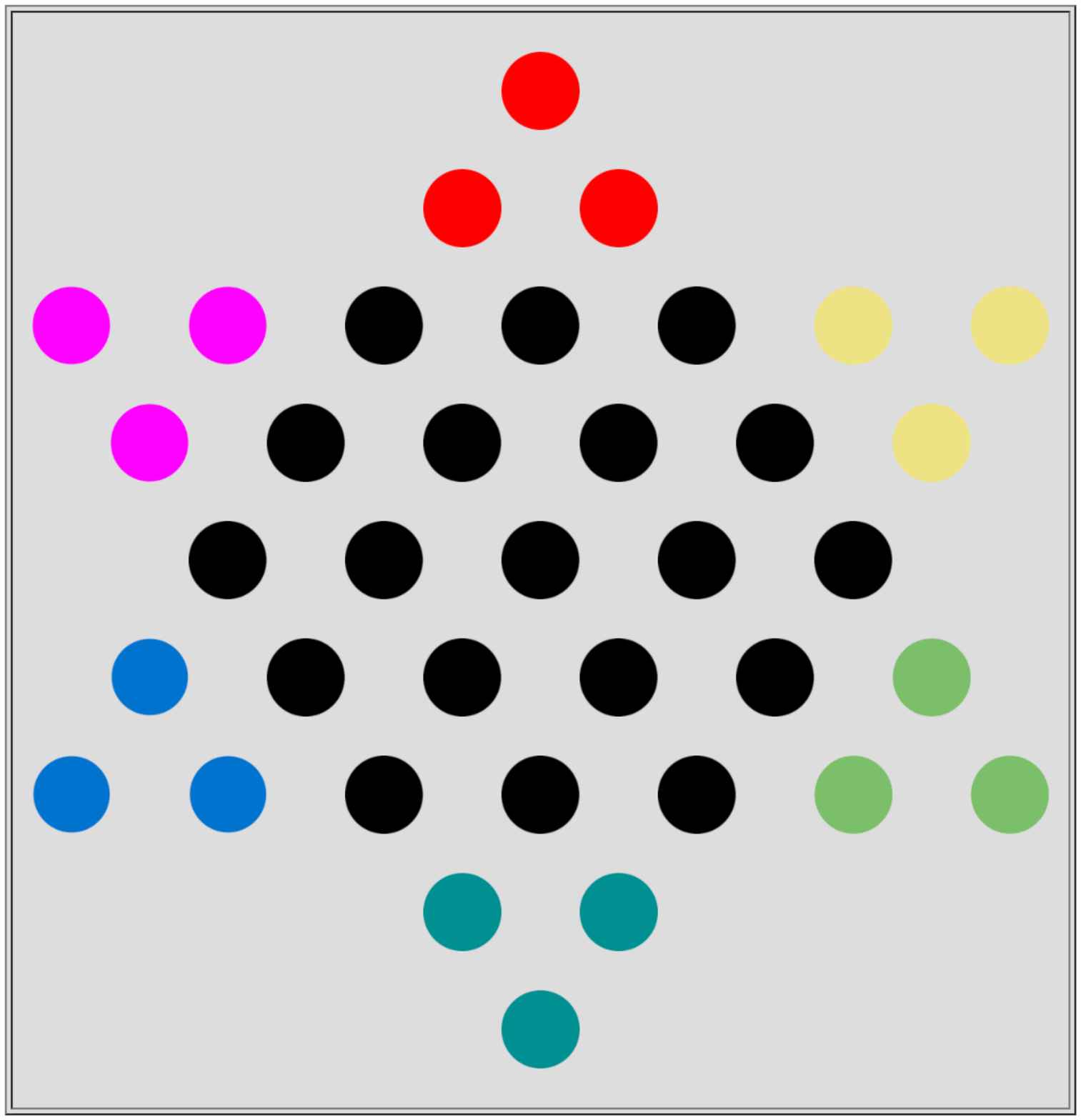} &
\includegraphics[scale=0.12,trim=0 250 0 250, clip]{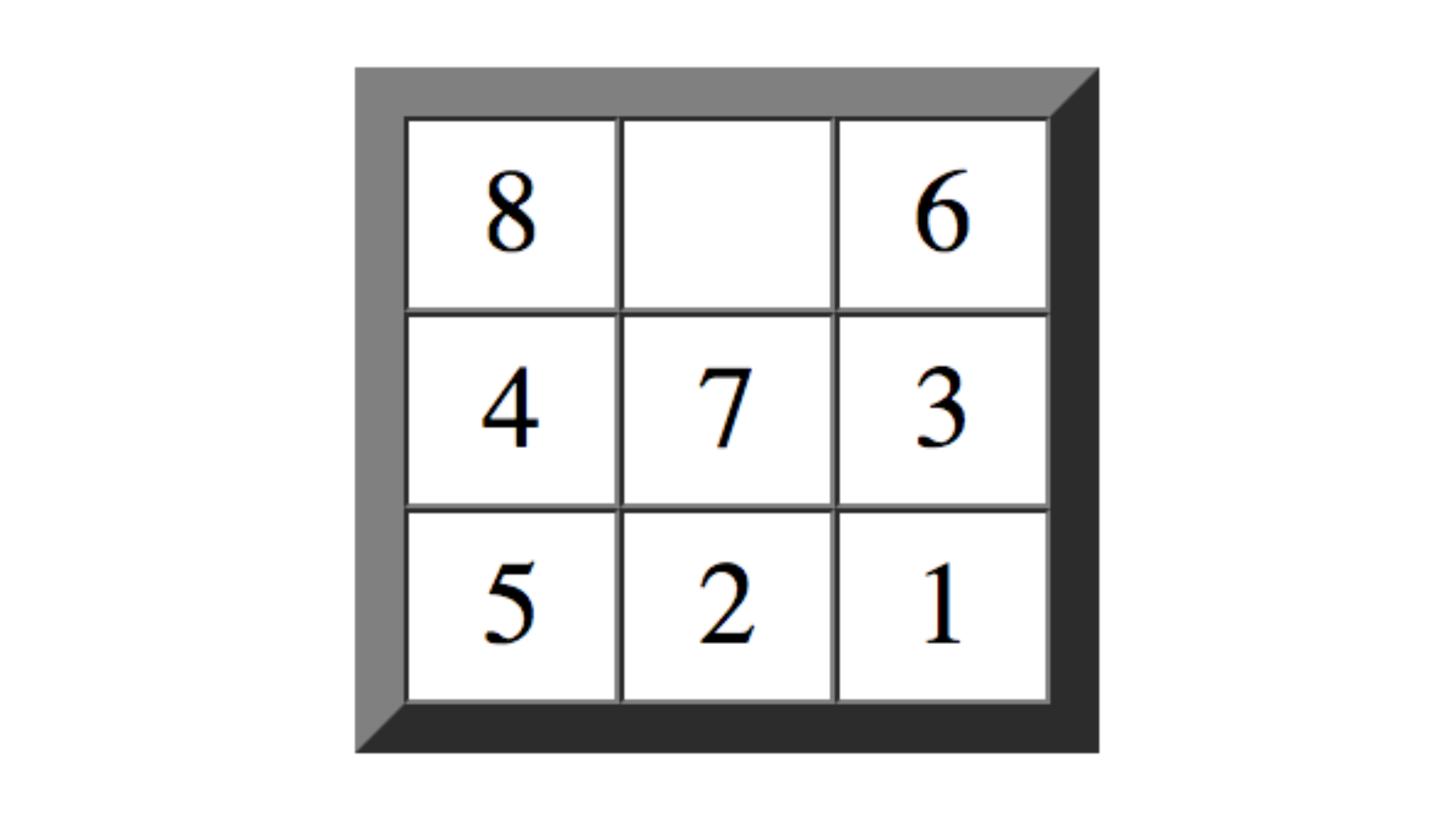} \\
\includegraphics[scale=0.12,trim=0 250 0 250, clip]{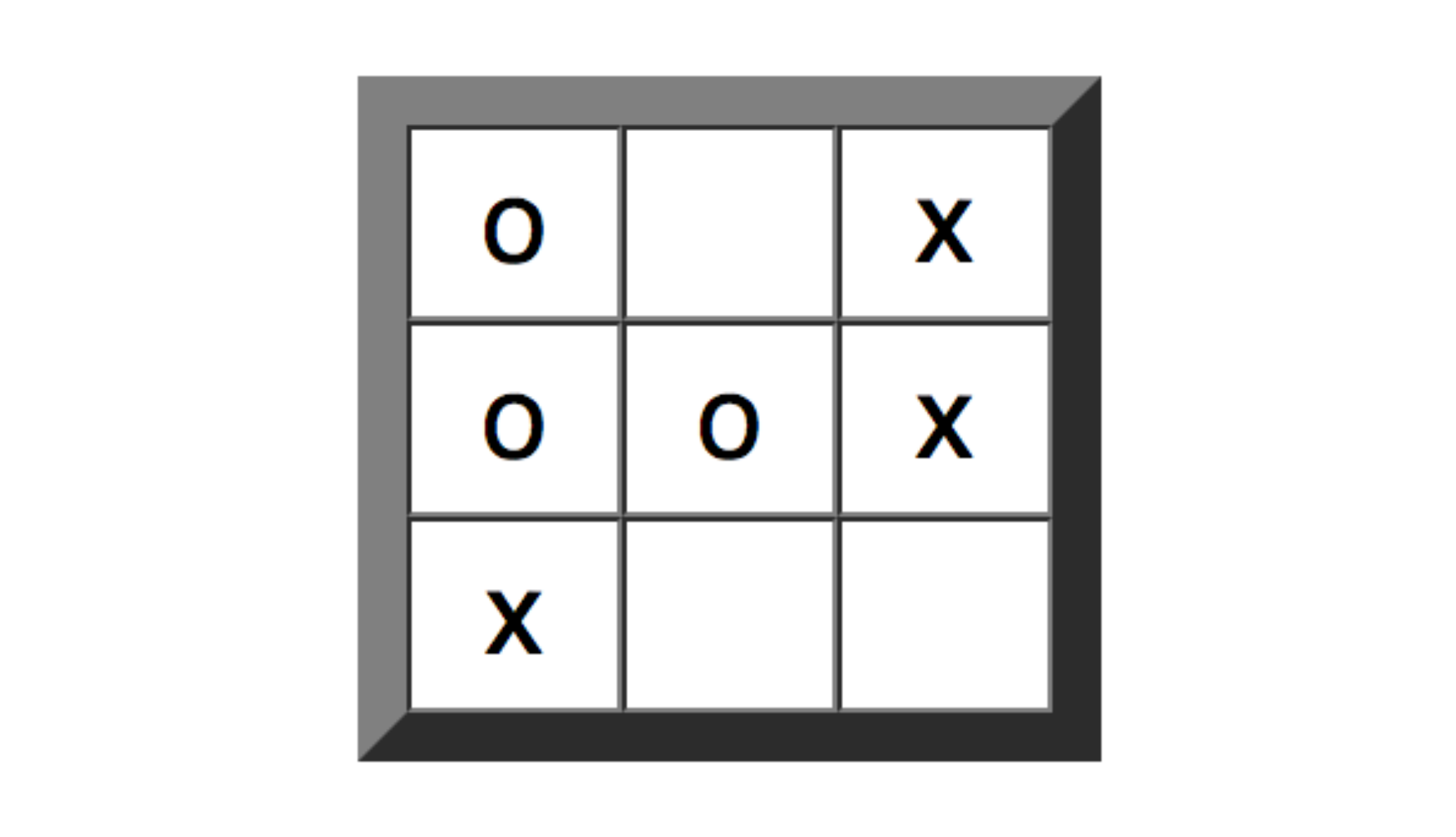} &
\includegraphics[scale=0.12,trim=0 100 0 100, clip]{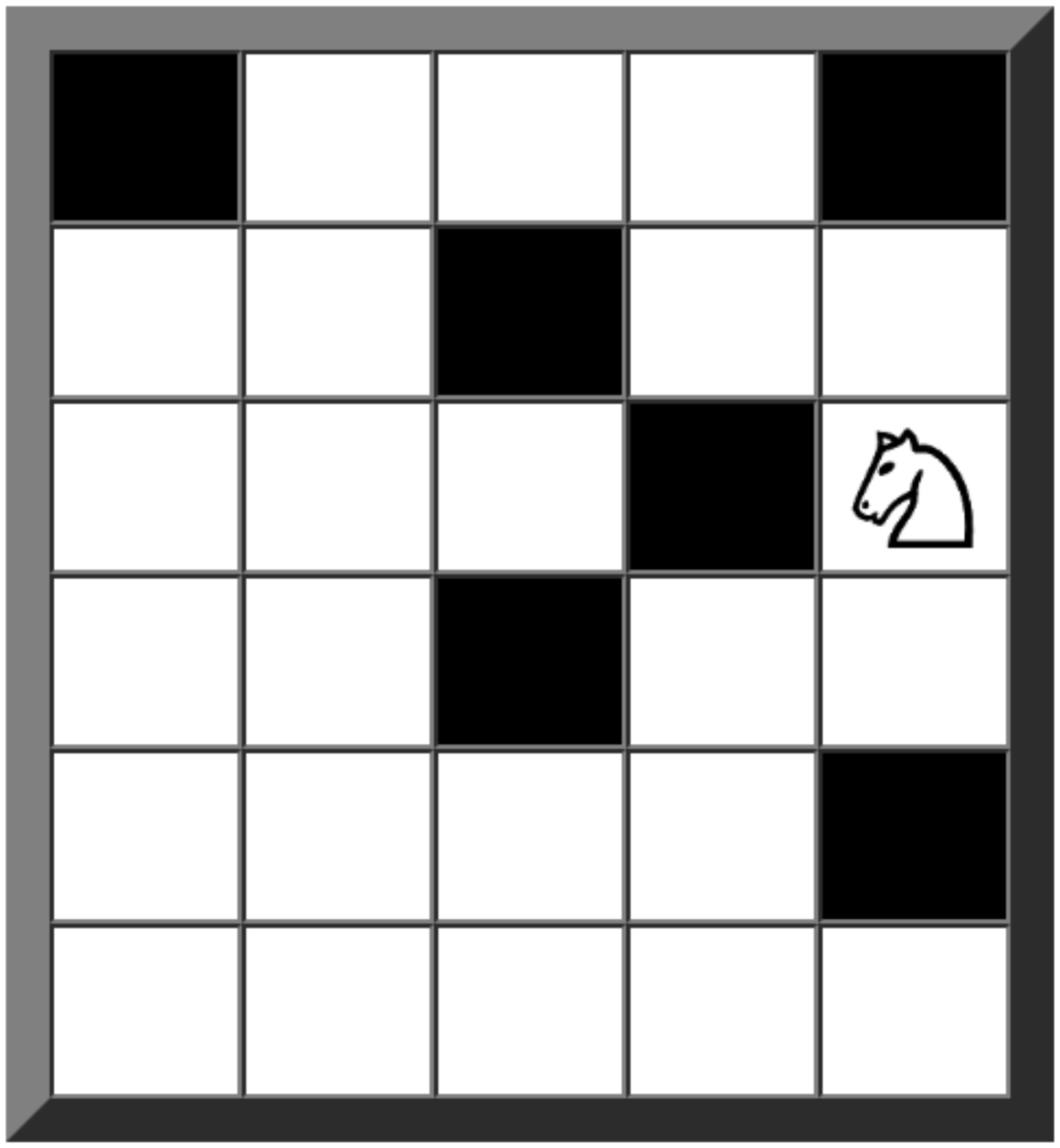} &
\includegraphics[scale=0.12,trim=0 100 0 100, clip]{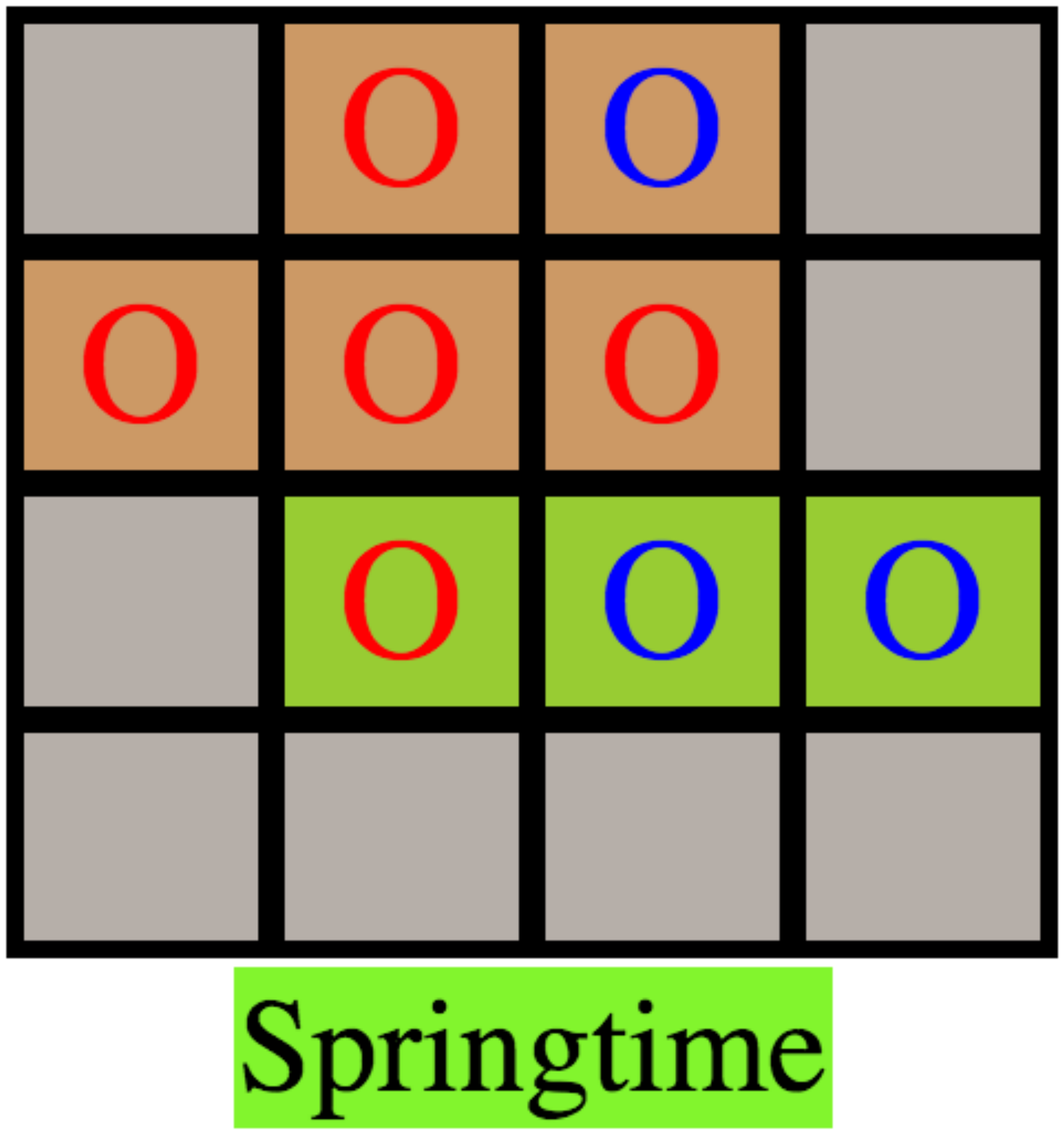}
\end{tabular}
\caption{Sample GGP games described in clockwise order starting from the top left: \emph{Alquerque}, \emph{Chinese Checkers}, \emph{Eight Puzzle}, \emph{Farming Quandries}, \emph{Knights Tour}, and \emph{Tic Tac Toe}.}
\label{fig:gamefigs}
\end{figure}

General game playing (GGP) \cite{genesereth} is a framework for evaluating an agent's general intelligence across a wide variety of games.
In the GGP competition, an agent is given the rules of a game that it has never seen before.
The rules are described in a first-order logic-based language called the game description language (GDL) \cite{love:gdl}.
The rules specify the initial game state, what constitutes legal moves, how moves update the game state, and how the game terminates \cite{bjornsson2012learning}.
Before the game begins, the agent is given a few seconds to think, to process the rules, and devise a game-specific strategy.
The agent then starts playing the game, thus generating game traces.
The winner of the competition is the agent that gets the best total score over all the games.
Figure \ref{fig:gamefigs} shows six example GGP games.
Figure \ref{fig:intro-rps-gdl} shows a selection of rules, written in GDL, for the game \emph{Rock Paper Scissors}.

\begin{figure}
\centering
\begin{tabular}{p{1\textwidth}}
\begin{minted}[frame=single]{scheme}
(succ 0  1)
(succ 1  2)
(succ 2  3)
(beats scissors paper)
(beats paper stone)
(beats stone scissors)
(<= (next (step ?n)) (true (step ?m)) (succ ?m ?n))
(<= (next (score ?p ?n)) (true (score ?p ?n)) (draws ?p))
(<= (next (score ?p ?n)) (true (score ?p ?n)) (loses ?p))
(<= (next (score ?p ?n)) (true (score ?p ?n2)) (succ ?n2 ?n) (wins ?p))
(<= (draws ?p) (does ?p ?a) (does ?q ?a) (distinct ?p ?q))
(<= (wins ?p) (does ?p ?a1) (does ?q ?a2) (distinct ?p ?q) (beats ?a1 ?a2))
(<= (loses ?p) (does ?p ?a1) (does ?q ?a2) (distinct ?p ?q) (beats ?a2 ?a1))
\end{minted}
\\
\end{tabular}
\caption{
A selection of rules for the game \emph{Rock Paper Scissors}.
The rules are written in the \emph{game description language}, a variant Datalog which is usually described in prefix notation.
The relation \tw{(succ 0  1)} means \tw{succ(0,1)}, i.e. 1 is the successor of 0.
Variables begin with "?".
The relation \mintinline{scheme}{(<= (next (step ?n)) (true (step ?m)) (succ ?m ?n))} can be rewritten in Prolog notation as \mintinline{prolog}{next(step(N)):- true(step(M)),succ(M,N).}
}
\label{fig:intro-rps-gdl}
\end{figure}

In this paper, we invert the GGP competition task: the learner (a machine learning system) is given game traces and the task is to induce (learn) the rules that could have produced the traces.
In other words, the learner must learn the rules of a game by observing others play.
This problem is a core part of \emph{inductive general game playing} (IGGP) \cite{Genesereth:aimag}, the task of jointly learning the rules of a game and playing the game successfully.
We focus exclusively on the first task.
Once the rules of the game have been learned then existing GGP techniques \cite{finnsson2012simulation,koriche2016stochastic,koriche2017woodstock} can be used to play the games.

Figure \ref{fig:rpc-example} shows an example IGGP task, described as a logic program, for the game \emph{Rock Paper Scissors}.
In this task, a learner is given a set of ground atoms representing background knowledge ($BK$) and sets of disjoint ground atoms representing positive ($E^+$) and negative ($E^-$) examples of target concepts.
The task is for the learner to induce a set of general rules (a logic program) that explains all of the positive but none of the negative examples.
In this scenario, the examples are observations of the \tw{next\_score} and \tw{next\_step} predicates, and the task is to learn the rules for these predicates, such as the rules shown in Figure \ref{fig:rpc-prog}.






\begin{figure}[ht]
\centering
\begin{tabular}{p{0.25\textwidth}|p{0.2\textwidth}|p{0.2\textwidth}}
$BK$ & $E^+$ & $E^-$\\
\hline
\begin{minted}{prolog}
beats(paper,stone).
beats(scissors,paper).
beats(stone,scissors).
player(p1).
player(p2).
succ(0,1).
succ(1,2).
succ(2,3).
does(p1,stone).
does(p2,paper).
true_score(p1,0).
true_score(p2,0).
true_step(0).
\end{minted}
&
\begin{minted}{prolog}
next_score(p1,0).
next_score(p2,1).
next_step(1).
\end{minted}
&
\begin{minted}{prolog}
next_score(p2,0).
next_score(p1,1).
next_score(p1,2).
next_score(p2,2).
next_score(p1,3).
next_score(p2,3).
next_step(0).
next_step(2).
next_step(3).
\end{minted}
\\
\end{tabular}
\caption{
An example learning task for the game \emph{Rock Paper Scissors}.
The input is a set of ground atoms representing background knowledge ($BK$) and sets of ground atoms representing positive ($E^+$) and negatives ($E^-$) examples.
In this task, the examples are observations of the \tw{next\_score} and \tw{next\_step} predicates.
The task is to learn the rules for these predicates, such as the rules shown in Figure \ref{fig:rpc-prog}.
}
\label{fig:rpc-example}
\end{figure}

\begin{figure}[ht]
\centering
\begin{tabular}{|p{0.25\textwidth}p{0.25\textwidth}|}
\hline
\begin{minted}{prolog}
next_step(N):-
    true_step(M),
    succ(M,N).
next_score(P,N):-
    true_score(P,N),
    draws(P).
next_score(P,N):-
    true_score(P,N),
    loses(P).
next_score(P,N2):-
    true_score(P,N1),
    succ(N2,N1),
    wins(P).
\end{minted}
&
\begin{minted}{prolog}
draws(P):-
    does(P,A),
    does(Q,A),
    distinct(P,Q).
loses(P):-
    does(P,A1),
    does(Q,A2),
    distinct(P,Q),
    beats(A2,A1).
wins(P):-
    does(P,A1),
    does(Q,A2),
    distinct(P,Q),
    beats(A1,A2).
\end{minted}
\\
\hline
\end{tabular}
\caption{
The GGP reference solution for the \emph{Rock Paper Scissors} game described as a logic program.
Note that the predicates \tw{draws}, \tw{loses}, and \tw{wins} are not given as background knowledge and the learner must discover these.
}
\label{fig:rpc-prog}
\end{figure}

In this paper, we expand on the idea proposed by Genesereth \cite{Genesereth:aimag} and we introduce the IGGP problem (Section \ref{sec:iggpprob}).
Our main claim is that IGGP is difficult for existing inductive logic programming (ILP) techniques, and in Section \ref{sec:rw} we outline the reasons why we think IGGP is difficult, such as the lack of task-specific language biases.
To support our claim, we make three key contributions.

Our main contribution is a new IGGP dataset\footnote{The dataset is available at https://github.com/andrewcropper/iggp}.
The dataset is based on game traces from 50 games from the GGP competition.
The games vary across a number of dimensions, including the number of players (1-4), the number of spatial dimensions (0-2), the reward structure (whether the rewards are zero-sum, cooperative, or orthogonal), and complexity.
Some of the games are turn-taking (\emph{Alquerque}) while others (\emph{Rock Paper Scissors}) are simultaneous.
Some of the games are classic board games (\emph{Checkers} and \emph{Hex}); some are puzzles (\emph{Sokoban} and \emph{Sudoku}); some are dilemmas from game theory (\emph{Prisonner's Dilemma} and \emph{Chicken}); others are simple implementations of classic video games (\emph{Centipede} and \emph{Tron}).
Figure \ref{fig:games} lists the 50 games and also shows for each game the number of dimensions, the number of players, and as an estimate of the game's complexity the number of rules and literals in the GGP reference solution.
Each game is described as four relational learning tasks \tw{goal}, \tw{next}, \tw{legal}, and \tw{terminal} with varying arities, although flattening the dataset to remove function symbols leads to more relations as illustrated in Figure \ref{fig:rpc-example} where the \tw{next} predicate is flattened to relations \tw{next\_score/2} and \tw{next\_step/2}.
For each game, we provide (1) training/validate/test data composed of sets of ground atoms in a 4:1:1 split, (2) a type signature file describing the arities of the predicates and types of the arguments, and (3) a reference solution in GDL.
It is important to note that we have not designed these games: the games were designed independently from our IGGP problem without this induction task in mind.

Our second contribution is a mechanism to continually expand the dataset.
The GGP competition produces new games each year, which provides a continual rich source of challenges to the GGP participants.
Our technical contribution allows us to easily add these new games to our dataset.
We implemented an automatic procedure for producing a new learning task from a game.
When a new game is added to the GGP competition, our system can read the GDL description, generate traces of sample play, and extract an IGGP task from those traces (see Section \ref{sec:generating-tasks} for technical details).
This automatic procedure means that our dataset can expand each year as new games are added to the GGP competition.
We again stress that the GGP games were not designed with this induction task in mind.
The games were designed to be challenging for GGP systems.
Thus, this induction task is based on a challenging ``real world'' problem, not a task that was designed to be the appropriate level of difficulty for current ILP systems.

Our third contribution is an empirical evaluation of existing ILP approaches, to test our claim that IGGP is difficult for current ILP approaches.
We evaluate the classical ILP system Aleph \cite{aleph} and the more recent systems ASPAL~\cite{aspal}, Metagol \cite{metagol}, and ILASP \cite{law:ilasp}.
Although non-exhaustive, these systems cover a breadth of ILP approaches and techniques.
We also compare non-ILP approaches in the form of simple baselines and clustering (KNN) approaches.
Figure \ref{tab:summary} summarises the results.
Although some systems can solve some of the simpler games, most of the games cannot be solved by existing approaches.
In terms of \emph{balanced accuracy} (Section \ref{sec:ba}), the best performing system, ILASP, achieves 86\%.
However, in terms of our \emph{perfectly solved} metric (Section \ref{sec:ps}), the best performing system, ILASP, achieves only 40\%.
Our empirical results suggest that our current IGGP dataset poses many challenges to existing ILP approaches.
Furthermore, because of our second contribution, our dataset will continue to grow with the GGP competition, as new games are added every year.
We therefore think that the IGGP problem and dataset will be valuable for motivating and evaluating future research.

\begin{figure}[!htb]
\small
    \begin{minipage}{.5\linewidth}
      \centering
        \begin{tabular}{|l|l|l|l|l|}
\hline
\textbf{Game} & \textbf{R} & \textbf{L} & \textbf{D} & \textbf{P} \\ \hline
Minimal Decay & 2 & 6 & 0 & 1 \\ \hline
Minimal Even & 8 & 19& 0 & 1 \\ \hline
Rainbow & 10 & 48& 0 & 1 \\ \hline
Rock Paper Scissors & 12 &36 & 0 & 1 \\ \hline
GT Chicken & 16&78 & 0 & 2 \\ \hline
GT Attrition & 16&60 & 0 & 2 \\ \hline
Coins & 16&45 & 0 & 1 \\ \hline
Buttons and Lights & 16&44 & 1 & 1 \\ \hline
Leafy & 17&80 & 2 & 2 \\ \hline
GT Prisoner & 17&75 & 0 & 2 \\ \hline
Eight Puzzle & 17&60 & 2 & 1 \\ \hline
Lightboard & 18&69 & 2 & 2 \\ \hline
Knights Tour & 18&46 & 2 & 1 \\ \hline
Sukoshi & 19&49 & 1 & 2 \\ \hline
Walkabout & 22&66 & 2 & 2 \\ \hline
Horseshoe & 22&59 & 2 & 2 \\ \hline
GT Ultimatum & 22&67 & 0 & 2 \\ \hline
Tron & 23&76 & 2 & 2 \\ \hline
9x Buttons and Lights & 24&77 & 2 & 1 \\ \hline
Hunter & 24&69 & 2 & 1 \\ \hline
GT Centipede & 24&69 & 0 & 2 \\ \hline
Fizz Buzz & 25&74 & 0 & 1 \\ \hline
Untwisty Corridor & 27&68 & 0 & 1 \\ \hline
Don't Touch & 29&84 & 2 & 2 \\ \hline
Tiger vs Dogs & 30&88 & 2 & 2 \\ \hline
\end{tabular}
    \end{minipage}%
    \begin{minipage}{.5\linewidth}
      \centering
        \begin{tabular}{|l|l|l|l|l|}
\hline
\textbf{Game} & \textbf{R} & \textbf{L} & \textbf{D} & \textbf{P} \\ \hline
Sheep and Wolf & 30&89 &2 &2 \\ \hline
Duikoshi & 31&76 & 2 & 2 \\ \hline
TicTacToe & 32&92 & 2 & 2 \\ \hline
HexForThree & 35&130 & 2 & 3 \\ \hline
Connect 4 & 36&124 & 2 & 4 \\ \hline
Breakthrough & 36&126 & 2 & 2 \\ \hline
Centipede & 37&134 & 2 & 1 \\ \hline
Forager & 40&106 & 2 & 1 \\ \hline
Sudoku & 41&101 & 2 & 1 \\ \hline
Sokoban & 41&172 & 2 & 1 \\ \hline
9x TicTacToe & 42&149 & 2 & 2 \\ \hline
Switches & 44&183 & 2 & 1 \\ \hline
Battle of Numbers & 44&134 & 2 & 2 \\ \hline
Free For All & 46&130 & 2 & 2 \\ \hline
Alquerque & 49&134 & 2 & 2 \\ \hline
Kono & 50&134 & 2 & 2 \\ \hline
Checkers & 52&167 & 2 & 2 \\ \hline
Pentago & 53&188 & 2 & 2 \\ \hline
Platform Jumpers & 62&168 & 2 & 2 \\ \hline
Pilgrimage & 80&240 & 2 & 2 \\ \hline
Firesheep & 85&290 & 2 & 2 \\ \hline
Farming Quandries & 88&451 & 2 & 2 \\ \hline
TTCC4 & 94&301 & 2 & 2 \\ \hline
Frogs and Toads & 97&431 & 2 & 2 \\ \hline
Asylum & 101&273 & 2 & 2 \\ \hline
\end{tabular}
    \end{minipage}
    \caption{The IGGP dataset. We list the number of rules (clauses) \textbf{R}, the number of literals \textbf{L}, number of dimensions \textbf{D}, and the number of players \textbf{P}.}
\label{fig:games}
\end{figure}

\begin{figure}[ht]
\centering
\begin{tabular}{|l|c|c|c|c|c|c|}
\hline
Metric & \textbf{Baseline} & \textbf{KNN$_5$} & \textbf{Aleph} & \textbf{ASPAL} & \textbf{Metagol}  & \textbf{ILASP$^{*}$} \\
\hline
  Balanced accuracy (\%) & 48 & 80 & 66 & 55 & 69 & \textbf{86}\\
\hline
  Perfectly solved (\%) & 4 & 19 & 18 & 10 & 34 & \textbf{40}\\
\hline
\end{tabular}
\caption{
Results summary.
The baseline represents accepting everything.
The results show that all of the approaches struggle in terms of the perfectly solved metric (which represents how many tasks were solved with 100\% accuracy).
}
\label{tab:summary}
\end{figure}

The rest of the paper is organised as follows.
Section 2 describes related work and further motivates this new problem and dataset.
Section 3 describes the IGGP problem, the GDL, in which GGP games are described, and how IGGP games are Markov games.
Section 4 introduces a technique to produce a IGGP task from a GGP game and provides specific details on how we generated our initial IGGP dataset.
Section 5 describes the baselines and ILP systems used in the evaluation of current ILP techniques.
Section 6 details the results of the evaluation and also describes why IGGP is so challenging for existing approaches.
Finally, Section 6 concludes the paper and details future work.

\section{Related work}
\label{sec:rw}

\subsection{General game playing}

As Björnsson states \cite{bjornsson2012learning}, from the inception of AI games have played a significant role as a test-bed for advancing the field.
Although the early focus was on developing general problem-solving approaches, the focus shifted towards developing problem-specific approaches,
such as approaches to play chess \cite{deepblue} or checkers \cite{chinook} very well.
One motivation of the GGP competition is to reverse this shift, as to encourage work on developing general AI approaches that can solve a variety of problems.

Our motivation for introducing the IGGP problem and dataset is similar.
As we will discuss in the next section, there is much work in ILP on learning rules for specific games, or for specific patterns in games.
However, there is little work on demonstrating general techniques for learning rules for a wide variety of games (i.e. the IGGP problem).
We want to encourage such work by showing that current ILP systems struggle on this problem.

\subsection{Inducing game rules}

Inducing game rules has a long history in ILP, where chess has often been the focus.
Bain~\cite{bain1994learning} studied inducing first-order Horn rules to determine the legality of moves in the chess KRK (king-rook-king) endgame, which is similar to the problem of learning the \tw{legal} predicate in the IGGP games.
Bain also studied inducing rules to optimally play the KRK endgame.
Other works on chess include Goodacre~\cite{goodacre1996inductive}, Morales~\cite{DBLP:journals/ci/Morales96}, who induced rules to play the KRK endgame and rules to describe the fork pattern, and Muggleton et al. \cite{chess-revision}.

Besides chess, Castillo and Wrobel~\cite{DBLP:conf/ijcai/CastilloW03} used a top-down ILP system and active learning to induce a rule for when a square is safe in the game minesweeper. Law et al. ~\cite{law:ilasp} used an ASP-based ILP approach to induce the rules for Sudoku and showed that this more expressive formalism allows for game rules to be expressed more compactly.

Kaiser~\cite{kaiser2012learning} learned the legal moves and the win condition (but not the state transition function) for a variety of boardgames (breakthrough, connect4, gomuku, pawn whopping, and tictactoe).
This system represents game rules as formulas of first-order logic augmented with a transitive closure operator $TC$; it learns by enumerative search, starting with the guarded fragment before proceeding to full first-order logic with $TC$.
Unusually, their system learns the game rules from \emph{videos} of correct and incorrect play: before it can start learning the rules, it has to parse the video, converting a sequence of pixel arrays into a sequence of sets of ground atoms.

Relatedly, Grohe and Ritzert ~\cite{grohe2017learning} also use enumerative search, searching through the space of first-order formulas.
They exploit Gaifman's locality theorem to search through a restricted set of local formulas.
They show, remarkably, that if the max degree of the Gaifman graph is polylogarithmic in the number $n$ of objects, then the running time of their enumerative learning algorithm is also polylogarithmic in $n$.
This intriguing result does not, however, suggest a practical algorithm as the constants involved are very large.

GRL~\cite{gregory2015grl} builds on SGRL~\cite{bjornsson2012learning} and LOCM~\cite{cresswell2009acquisition} to learn game dynamics from traces.
In these systems, the game dynamics are modelled as as finite deterministic automata.
They do not learn the \tw{legal} predicate (determining which subset of the possible moves are available in the current state) or the \tw{goal} predicate.

As is clear from these works, there is little work in ILP demonstrating general techniques for learning rules for a wide variety of games.
This limitation partially motivates the introduction of the IGGP problem and dataset.

\subsection{Existing datasets}

One of our main contributions is the introduction of a IGGP dataset.
In contrast to the existing datasets, our dataset introduces many new challenges.

\subsubsection{Size and diversity}

Our dataset is larger and more diverse than most existing ILP datasets, especially on learning game rules.
Commonly used ILP datasets, such as kinship data \cite{hinton1986learning}, Michaslki trains \cite{michalski:trains}, Mutagenesis \cite{mutagenesis}, Carcinogenesis \cite{carcinogenesis}, string transformations \cite{mugg:metabias}, and chess positions \cite{chess-dataset}, typically contain a single predicate to be learned, such as \tw{eastbound/1} or \tw{westbound/1} in the Michaslki trains dataset or \tw{active/1} in the Mutagenesis dataset.
By contrast, our dataset contains 50 distinct games, each described by at least four target predicates, where flattening leads to more relations as illustrated in Figure \ref{fig:rpc-example}.
In addition, whereas some datasets use only dyadic concepts, such as kinship or string transformations, our dataset also requires learning programs with a mixture of predicates arities, such as \tw{input\_jump/8} in \emph{Checkers} and \tw{next\_cell/4} predicate in \emph{Sudoku}.
Learning programs with high-arity predicates is a challenge for some ILP approaches ~\cite{metagol,hexmil,evans:dilp}.
Moreover, because of our second main contribution, we can continually and automatically expand the dataset as new games are introduced into the GGP competition.
Therefore, our IGGP dataset will continue to expand to include more games.

\subsubsection{Inductive bias}
Our IGGP games come from the GGP competition.
As stated in the introduction, the games were not designed with this induction task in mind.
One key challenge proposed by the IGGP problem is the lack of inductive bias provided.
Most existing work on inducing game rules has assumed as input a set of high-level concepts.
For instance, Morales~\cite{DBLP:journals/ci/Morales96} assumed as input a predicate to determine when a chess piece is in check.
Likewise, Law \cite{law:ilasp} assumed high-level concepts such as \tw{same\_row/2} and \tw{same\_col/2} as background knowledge when learning whether a \emph{Sudoku} board was valid.
Moreover, most existing ILP work on game learning rules (and learning in general) involves the designers of the system designing the appropriate representation of the problem for their system.
By contrast, in our IGGP problem the representation is fixed: it is the GDL provided by the GGP.

Many existing ILP techniques assume a task-specific language bias, expressing a hypothesis space which contains at least one correct representation of the target concept.
When available, language biases are extremely useful as a smaller hypothesis space can mean fewer examples and less computational resources are needed by the ILP systems.
In many practical situations, however, task-specific language biases are either not available, or are extremely wide, as very little information is known about the structure of the target concept.

In our IGGP dataset we only provide the most simple (or primitive) low-level concepts, which come directly from the GGP competition, i.e. our IGGP dataset does not provide any task-specific language biases.
For each game, the only language bias given is the type schema of each predicate in the language of the background knowledge.
For instance, in \emph{Sudoku} the higher-level concepts of \emph{same row} and \emph{same col} are not given.
Likewise, to learn the \tw{terminal} predicate in \emph{Connect Four}, a learner must learn the concept of a line, which in turn requires learning rules for vertical, horizontal, and diagonal lines.
This means that for an approach to solve the IGGP problem in general (and to be able to accept future games without changing their method), it must be able to learn without a game-specific bias, or be able to generate this game-specific bias from the type-schemas in the task.
In addition, a learner must learn concepts from only primitive low-level background predicates, such as \tw{cell(X,Y,Filled)}.
Should these high-level concepts be reusable then it would be advantageous to perform predicate invention, which has long been a key challenge in ILP \cite{mlj:ilp20,mugg:metalearn}.
Popular ILP systems, such as FOIL~\cite{foil} and Progol~\cite{mugg:progol}, do not support predicate invention, and although recent work~\cite{inoue:mla,mugg:metagold,crop:metafunc} has tackled this challenge, predicate invention is still a difficult problem.

\subsubsection{Large programs}
Many reference solutions for IGGP games are large, both in terms of the number of literals and the clauses in them.
For instance, the GGP reference solution for the \tw{goal} predicate for \emph{Connect Four} uses 14 clauses and a total of 72 literals.
This solution uses predicate invention to essentially compress the solution, where the auxillary predicates include the concept of a line, which in turn uses the auxillary predicates for the concepts of columns, rows, and diagonals.
If we unfold the reference solution as to remove auxillary predicates then the total number of literals required to learn a solution for this single predicate easily exceeds 400.
However, learning large programs is a challenge for most ILP systems~\cite{crop:thesis} which typically struggle to learn programs with hundreds of clauses or literals.

\subsubsection{ILP2016 competition}
\label{sec:ilp2016}
The closest work similar to ours is the ILP 2016 Competition \cite{ILP16Comp}.
The ILP 2016 competition was based on a single type of task (with various hand crafted target hypotheses) aimed at learning the valid moves of an agent as it moved through a grid.
In some ways this is similar to our \tw{legal} tasks, although many tasks required learning invented predicates representing changes in state, similar to our \tw{next} tasks.
By contrast, our IGGP problem and dataset is based on a variety of real games, which we did not design.
Furthermore, the ILP 2016 dataset provides restricted inductive biases to aid the ILP systems, whereas we (deliberately) do not give such help.





\subsection{Model learning}

AlphaZero \cite{silver2017mastering} has shown the power of combining tree search with a deep neural network for distilling search policy into a neural net.
But this technique presupposes that we have been given a model of the game dynamics: we must already know the state transition function and the reward function.
Suppose we want to extend AlphaZero-style techniques to domains where we are \emph{not} given an explicit model of the environment.
We would need some way of learning a model of the environment from traces.
Ideally, we would like to learn data-efficiently, without needing hundreds of thousands of traces.

Model-free reinforcement learning agents have high sample complexity: they often require millions of episodes before they can learn a reasonable policy.
Model-based agents, by contrast, are able to use their understanding of the dynamics of the environment to learn much more efficiently~\cite{dvzeroski,duff,guez}.
Whether, and to what extent, model-based methods are more sample efficient than model-free methods depends on the complexity of the particular MDP.
Sometimes, in simple environments, one needs fewer data to learn a policy than to learn a model.
It has also been shown that, for Q learning, the worst-case asymptotics for model-based and model-free are the same \cite{kearns1999finite}.
But these qualifications do not, of course, undermine the claim that in complex environments that require anticipation or planning, a model-based agent will be significantly more sample-efficient than its model-free counterpart.

The GGP dataset was designed to test an agent's ability to learn a model that can be useful in planning.
The most successful GGP algorithms, e.g. Cadiaplayer~\cite{finnsson2012simulation}, Sancho \cite{koriche2016stochastic}, and WoodStock~\cite{koriche2017woodstock}, use Monte Carlo Tree Search (MCTS) to search.
MCTS relies on an accurate forward model of the Markov Decision Process.
The further into the future we search, the more important it is that our forward model is accurate, as errors compound.
In order to avoid having to give our MCTS agents a hand-coded model of the game dynamics,
they must be able to learn an accurate model of the dynamics from a handful of behavior traces.

Two things make the GGP dataset an appealing task for model learning. First, hundreds of games have already been designed for the GGP competition, with more being added each year. Second, each game comes with `ground truth': a set of rules that completely describe the game. From these rules, we know the learning problem is solvable, and we have a good measure of how hard it is (by measuring the complexity of the ground-truth program\footnote{This measure of complexity assumes, of course, that the length of the ground-truth program is reasonably close to the shortest GDL description of the game. In other words, this assumes the actual program length is a reasonable estimate of the Kolmogorov complexity.}).



\section{IGGP dataset}
\label{sec:dataset}

In this section, we describe the Game Description Language (GDL) in which GGP games are described, the IGGP problem setting, and finally an illustrative example of a typical IGGP task.

\subsection{Game description language}
\label{sec:gdl}


GGP games are described using GDL.
This language describes the state of a game as a set of facts and the game mechanics as logical rules.
GDL is a variant of Datalog with two syntactic extensions (stratified negation and restricted function symbols) and with a small set of distinguished predicates that have a special meaning \cite{love:gdl} (shown in Figure \ref{tab:gdl}).

The first syntactic extension is stratified negation.
Standard Datalog (lacking negation altogether) has the useful property that there is a unique minimal model \cite{dantsin2001complexity}.
If we add unrestricted negation, we lose this attractive property: now there can be multiple distinct minimal models.
To maintain the property of having a unique minimal model, GDL adds a restricted form of negation called \emph{stratified} negation \cite{stratnegation}.
The \emph{dependency graph} of a set of rules is formed by creating an edge from predicate $p$ to predicate $q$ whenever there is a rule whose head is $p(...)$ and that contains an atom $q(...)$ in the body.
The edge is labelled with a negation if the body atom is negated.
A set of rules is \emph{stratified} if the dependency graph contains no cycle that includes a negated edge.

GDL's second syntactic extension to Datalog is restricted function symbols.
The Herbrand base of a standard Datalog program is always finite.
If we add unrestricted function symbols, the Herbrand base can be infinite.
To maintain the property of having a finite Herbrand base, GDL restricts the use of function symbols in recursive rules \cite{love:gdl}.

The two syntactic extensions of GDL, stratified negation and restricted function symbols, mean we extend the expressive power of Datalog without essentially changing its key attractive property: there is always a single, finite minimal model \cite{love:gdl}.


\begin{figure}[ht]
\begin{tabular}{l|l}
\textbf{Predicate} & \textbf{Description} \\
\hline
\tw{distinct(?x,?y)} & Two terms are syntactically different\\
\tw{does(?r,?m)} & Player \tw{?r} performs action \tw{?m} in the current game state\\
\tw{goal(?r,?n)} & Player \tw{?r} has reward \tw{?n} (usually a natural number) in the current state\\
\tw{init(?f)} & Atom \tw{?f} is true in the initial game state\\
\tw{legal(?r,?m)} & Action \tw{?m} is a legal move for player \tw{?r} in the current state\\
\tw{next(?f)} & Atom \tw{?f} will be true in the next game state\\
\tw{role(?n)} & Constant \tw{?n} denotes a player\\
\tw{terminal} & The current state is terminal\\
\tw{true(?f)} & Atom \tw{?f} is true in the current game state
\end{tabular}
\caption{Main predicates in GDL where variables begin with a "?" symbol.}
\label{tab:gdl}
\end{figure}

\subsection{Problem setting}
\label{sec:iggpprob}


We now define the IGGP problem.
Our problem setting is based on the ILP learning from entailment setting \cite{luc:book}, where an example corresponds to an observation about the truth or falsity of a formula $F$ and a hypothesis $H$ covers $F$ if $H$ entails $F$.
We assume languages of background knowledge $\mathcal{B}$ and examples $\mathcal{E}$ each formed of function-free ground atoms.
The atoms are function-free because we flatten the GDL atoms.
For example, in Figure \ref{fig:fb-legal}, the atom \tw{true(count(9))} has been flattened into \tw{true\_count(p9)}.
We flatten atoms because some ILP systems do not support function symbols.
We likewise assume a language of hypotheses $\mathcal{H}$ formed of datalog programs with stratified negation.
Stratified negation is not necessary but in practice allows significantly more concise programs, and thus often makes the learning task computationally easier.
Note that the GDL also supports recursion but in practice most GGP games do not use recursion.
In future work we intend to contribute recursive games to the GGP competition.

We now define the IGGP input:

\begin{definition}[\textbf{IGGP input}]
An \emph{IGGP input} $\Delta$ is a set of $m$ triples $\{(B_i,E^+_i,E^-_i)\}^m_{i=1}$ where
\begin{itemize}
\item $B_i \subset \mathcal{B}$ represents background knowledge
\item $E_i^+\subseteq \mathcal{E}$ and $E_i^-\subseteq \mathcal{E}$ represent positive and negative examples respectively
\end{itemize}
\end{definition}

\noindent
An IGGP input forms the \emph{IGGP problem}:

\begin{definition}[\textbf{IGGP problem}]
Given an IGGP input $\Delta$, the \emph{IGGP problem} is to return a hypothesis $H \in \mathcal{H}$ such that $\text{for all} \;\; (B_i,E^+_i,E^-_i) \in \Delta$ it holds that $H \cup B_i \models E^+$ and $H \cup B_i \not\models E_i^-$.
\end{definition}

\noindent
Note that a single hypothesis should be consistent with all given triples.

\subsubsection{Illustrating example: Fizz Buzz}

To give the reader an intuition for the IGGP problem and the GGP games, we now describe example scenarios for the game \emph{Fizz Buzz}.
Although typically a multi-player game, in our IGGP dataset \emph{Fizz Buzz} is a single-player game.
The aim of the game is for the player to replace any number divisible by three with the word \emph{fizz}, any number divisible by five with the word \emph{buzz}, and any number divisible by both three and five with \emph{fizzbuzz}.
For example, a game of Fizz Buzz up to the number 17 would go: 1, 2, fizz, 4, buzz, fizz, 7, 8, fizz, buzz, 11, fizz, 13, 14, fizz buzz, 16, 17.

Figures \ref{fig:fb-legal}, \ref{fig:fb-next}, \ref{fig:fb-goal}, and \ref{fig:fb-terminal} show example IGGP problems and solutions for the target predicates \tw{legal}, \tw{next}, \tw{goal}, and \tw{terminal} respectively.
For simplicity each example is a single $(B,E^+,E^-)$ triple, although in the dataset each learning task is often a set of multiple triples, where a single hypothesis should explain all the triples.
In all cases the BK shown in Figure \ref{fig:fb-bk} holds, so we omit it from the individual examples for brevity.
Note that the game only runs to the number 31.

\begin{figure}[ht]
\centering
\begin{tabular}{|p{0.35\textwidth}p{0.25\textwidth}|}
\hline
\begin{minted}{prolog}
divisible(12,1).
divisible(12,2).
...
divisible(12,12).
input_say(player,1).
input_say(player,2).
...
input_say(player,30).
input_say(player,fizz).
input_say(player,buzz).
input_say(player,fizzbuzz).
role(player).
int(0).
int(1).
...
int(31).
\end{minted}

&
\begin{minted}{prolog}
less_than(0,1).
less_than(0,2).
...
less_than(30, 31).
minus(1,1,0).
minus(2,1,1).
...
minus(31,31,0).
positive_int(1).
positive_int(2).
...
positive_int(31).
succ(0,1).
succ(0,2).
...
succ(30,31).
\end{minted}
\\
\hline
\end{tabular}
\caption{Common BK for \emph{Fizz Buzz}.}
\label{fig:fb-bk}

\end{figure}

\begin{figure}[ht]
\small
\centering
\begin{tabular}{p{0.17\textwidth}|p{0.3\textwidth}|p{0.23\textwidth}|p{0.25\textwidth}}
B & $E^+$ & $E^-$ & $H$\\
\hline

\begin{minted}{prolog}
true_count(9).
true_success(6).
\end{minted}

&
\begin{minted}{prolog}
legal_say(player,9).
legal_say(player,buzz).
legal_say(player,fizz).
legal_say(player,fizzbuzz).
\end{minted}
&
\begin{minted}{prolog}
legal_say(player,0).
legal_say(player,1).
...
legal_say(player,8).
legal_say(player,10).
...
legal_say(player,31).
\end{minted}
&
\begin{minted}{prolog}
legal_say(player,N):-
    true_count(N).
legal_say(player,fizz).
legal_say(player,buzz).
legal_say(player,fizzbuzz).
\end{minted}
\end{tabular}

\caption{In this \emph{Fizz Buzz} scenario the learner is given four positive examples of the \tw{legal\_say/2} predicate and many negative examples.
This predicate represents what legal moves a player can make in the game.
The column $H$ shows the reference GGP solution described as a logic program.
In \emph{Fizz Buzz}, the player can always make three legal moves in any state, saying fizz, buzz, or fizzbuzz.
The player can additionally say the current number (the counter).
}
\label{fig:fb-legal}
\end{figure}

\begin{figure}[ht]
\small
\centering
\begin{tabular}{p{0.25\textwidth}|p{0.18\textwidth}|p{0.18\textwidth}|p{0.25\textwidth}}
B & $E^+$ & $E^-$ & $H$\\
\hline

\begin{minted}{prolog}
does_say(player,buzz).
true_count(12).
true_success(3).
\end{minted}

&
\begin{minted}{prolog}
next_count(13).
next_success(3).
\end{minted}
&
\begin{minted}{prolog}
next_count(0).
next_count(1).
...
next_count(12).
next_count(14).
...
next_count(31).
next_success(0).
next_success(1).
next_success(2).
next_success(4).
...
next_success(31).
\end{minted}
&
\begin{minted}{prolog}
next_count(After):-
    true_count(Before),
    succ(Before,after).
next_success(After):-
    correct,
    true_success(Before),
    succ(Before,After).
next_success(A):-
    \+ correct,
    true_success(A).
correct:-
    true_count(N),
    divisible(N,15),
    does_player_say(fizzbuzz).
correct:-
    true_count(N),
    divisible(N,3),
    \+ divisible(N,5),
    does_player_say(fizz).
correct:-
    true_count(N),
    divisible(N,5),
    \+ divisible(N,3),
    does_player_say(buzz).
correct:-
    true_count(N),
    \+ divisible(N,5),
    \+ divisible(N,3),
    does_player_say(N).
\end{minted}
\end{tabular}
\caption{
In this \emph{Fizz Buzz} scenario, the learner is given one positive example of the \tw{next\_count/1} predicate, one positive example of the \tw{next\_success/1} predicate, and many negative examples of both predicates.
These predicates represent the change of game state.
The column $H$ shows the reference GGP solution described as a logic program, which may not necessarily be the most textually compact solution.
The \tw{next\_count/1} relation represents the count in the game.
The this relation has a single clause two literal definition, which says that the count increases by one after each step in the game.
The \tw{next\_success/1} relation requires two clauses with many literals.
This relation counts how many times a player says the correct output.
The reference GGP solution for this relation includes the \tw{correct/0} predicate which is not provided as BK but which is reused in both clauses of \tw{next\_success/1}.
For an ILP system to learn the reference solution it would need to invent this predicate.
Also note that this solution uses negation in the body, including the negation of the invented predicate \tw{correct/0}.
}
\label{fig:fb-next}
\end{figure}

\begin{figure}[ht]
\small
\centering
\begin{tabular}{p{0.20\textwidth}|p{0.2\textwidth}|p{0.20\textwidth}|p{0.2\textwidth}}
B & $E^+$ & $E^-$ & $H$\\
\hline

\begin{minted}{prolog}
true_count(31).
true_success(20).
\end{minted}

&

\begin{minted}{prolog}
goal(player,50).
\end{minted}

&
\begin{minted}{prolog}
goal(player,0).
goal(player,100).
goal(player,25).
goal(player,75).
\end{minted}

&
\begin{minted}{prolog}
goal(player,100):-
    true_success(30).
goal(player,75):-
    true_success(S),
    less_than(S,30),
    less_than(24,S).
goal(player,50):-
    true_success(S),
    less_than(S,25),
    less_than(19,S).
goal(player,25):-
    true_success(S),
    less_than(S,20),
    less_than(14,S).
goal(player,0):-
    true_success(S),
    less_than(S,15).
\end{minted}

\end{tabular}

\caption{
In this \emph{Fizz Buzz} scenario the learner is given one example of the \tw{goal/2} predicate and four negative examples.
This predicate represents the reward for a move.
In \emph{Fizz Buzz} the reward is based on the value of \tw{true\_success/1}.
The column $H$ shows the reference GGP solution described as a logic program.
The reference solution requires five clauses, which means that it would be difficult for ILP systems that only support learning single-clause programs \cite{mugg:progol,foil}.
}
\label{fig:fb-goal}
\end{figure}

\begin{figure}
\centering
\begin{tabular}{p{0.20\textwidth}|p{0.1\textwidth}|p{0.2\textwidth}|p{0.2\textwidth}}
B & $E^+$ & $E^-$ & $H$\\
\hline

\begin{minted}{prolog}
true_count(27).
true_success(8).
\end{minted}

&

\begin{minted}{prolog}
\end{minted}

&
\begin{minted}{prolog}
terminal.
\end{minted}

&
\begin{minted}{prolog}
terminal:-
    true_count(31).
\end{minted}

\end{tabular}

\caption{
In this \emph{Fizz Buzz} scenario the learner is given a single negative example of the \tw{terminal/0} predicate.
This predicate indicates when the game has finished.
In this scenario the game has not terminated.
In the dataset the \emph{Fizz Buzz} game runs until the count is 31, so the learner must learn a rule such as the one shown in column $H$.
}

\label{fig:fb-terminal}
\end{figure}

\section{Generating the GGP Dataset}

In this section, we describe our procedure to automatically generate IGGP tasks from GGP game descriptions.
We first explain how GGP games fit inside the framework of multi-agent Markov decision processes.
We also explain the need for a type-signature for each game.

\subsection{Preliminaries: Markov games}
\label{sec:markov-games}

GGP games are Markov games \cite{littman1994markov}, a strict superset of multi-agent Markov decision process (MDP)s that allow simultaneous moves\footnote{There are variants in which some games are stochastic, and some have imperfect information. But in the core GGP framework all games are deterministic and have perfect information.}. The four components $(S,A,T,R)$ of the MDP are:

\begin{itemize}
    \item $S$ is a finite set of states
    \item $A$ is a finite set of actions
    \item $T$ is transition function $T: S \times A \rightarrow S$
    \item $R$ is a reward function
\end{itemize}

\noindent
We describe these elements in turn for a GGP game.

\subsubsection{States}

Each state $s \in S$ is a set of ground atoms representing \emph{fluents} (propositions whose truth-value can change from one state to another). The \tw{true} predicate indicates which fluents are true in the current state. For instance, one state of a best-of-three game of \emph{Rock Paper Scissors} is:

\begin{center}
  \begin{tabular}{l}
    \tw{true(score(p1,0)).}\\
    \tw{true(score(p2,2)).}\\
    \tw{true(step(2)).}\\
  \end{tabular}
\end{center}

\noindent
This state represents that the current score is 0 to 2 in favour of player \tw{p2}, and 2 time-steps have been performed.

\subsubsection{Actions}

Each action $a \in A$ is a set of ground atoms representing the set of all joint actions for agents $1 .. n$. The \tw{does} predicate indicates which agents perform which actions. For instance, one set of joint actions for \emph{Rock Paper Scissors} is:

\begin{center}
  \begin{tabular}{l}
    \tw{does(p1,paper).}\\
    \tw{does(p2,stone).}\\
  \end{tabular}
\end{center}

\subsubsection{Transition function}
In a stochastic MDP, the transition function $T$ has the signature $T : S \times A \times S \rightarrow \{0,1\}$. By contrast, in a \emph{deterministic} MDP, such as a GGP game, the transition function is $T : S \times A \rightarrow S$. Given a current state $s$ and a set of actions $a$, the \tw{next} predicate indicates which fluents are true in the (unique) next state $s'$. For instance, in \emph{Rock Paper Scissors}, given the current state $s$ and actions $a$ above, the next state $s'$ is:

\begin{center}
  \begin{tabular}{l}
    \tw{next(score(p1,1)).}\\
    \tw{next(score(p2,2)).}\\
    \tw{next(step(3)).}\\
  \end{tabular}
\end{center}

\noindent
The transition function is a set of definite clauses defining \tw{next} in terms of \tw{true}. For instance, the following two clauses define part of the transition function for \emph{Rock Paper Scissors}:

\begin{center}
\begin{minipage}{4cm}
\begin{verbatim}
next(score(P,N2)):-
    does(P,paper),
    does(Q,stone),
    true(score(P,N1)),
    succ(N1,N2).
next(step(N2)):-
    true(step(N1)),
    succ(N1,N2).
\end{verbatim}
\end{minipage}
\end{center}


\subsubsection{Reward function}
In a continuous multi-agent MDP, the reward function has the signature\footnote{Sometimes, alternatively, the reward function has the slightly more expressive form  $R : S \times A \times S \rightarrow \mathbb{R}^n$.} $R : S \rightarrow \mathbb{R}^n$. In a discrete MDP, such as a GGP game, we assume a small fixed set of $k$ discrete rewards $\{r_1,\dots,r_k\}$, where $r_i$ is not necessarily numeric. Let $G[i]$ be the set of atoms representing that player $i$ has one of the $k$ rewards $G[i] = \{ goal(i, r_j) \mid j = 1 .. k \}$. Let $G = G[1] \times ... \times G[n]$ be the joint rewards for agents $1 .. n$. In our GGP dataset, the reward function has the signature $R : S \rightarrow G$. Note that, in this framework, learning the reward function becomes a \emph{classification} problem rather than a regression problem.
For example, in the \emph{Rock Paper Scissors} state above, the reward for state $s'$ depends only on the score and is:

\begin{center}
  \begin{tabular}{l}
    \tw{goal(p1,1).}\\
    \tw{goal(p2,2).}\\
  \end{tabular}
\end{center}

\subsubsection{Legal}
In the GGP framework, actions are sometimes unavailable.
It is not the case that all possible actions from $A$ can be performed, but some of them have no effect -- but rather that only a subset of actions are available in a particular state.

The legal function $L$ determines which actions are available in which states: $L : S \rightarrow 2^A$.
Recall that an element of $A$ is not an individual action performed by a single player, but rather a set of simultaneous joint actions, one for each player.
For example, one element of $A$ is $\{\tw{does(p1,paper).},  \tw{does(p2,stone).}\}$.
Note that the availability of an action for one agent does not depend on what other actions are being performed concurrently by other agents; it only depends on the state $S$.

\subsubsection{Terminal}
The GDL language contains a distinguished predicate, the nullary \tw{terminal} predicate,  that indicates when an episode has terminated (i.e. when the game is over).

\subsection{Preliminaries: the type-signature for a GGP game}
\label{sec:type-signature}

In order to calculate the complete set of ground atoms for a game\footnote{We could dispense with the type signature, and generate all possible untyped ground atoms. Naively generating all possible untyped ground atoms would significantly increase the size of the dataset. We use the type signature as a space optimisation to keep the dataset manageable.}, we use a type signature $\Sigma$.
The type signature defines the types of constants, functions, and predicates used in the GDL description.
Our type signatures include a simple subtyping mechanism for inclusion polymorphism. For example:
\begin{verbatim}
true, next :: prop -> bool.
at :: pos -> pos -> cell -> prop.
red, black :: agent.
1, 2, 3, 4, 5 :: pos.
blank :: cell.
agent :> cell.
\end{verbatim}
In this example, \verb|true| and \verb|next| are predicates, \verb|at| is a function that takes an $(x,y)$ coordinate and a cell-type and returns a fluent (\verb|prop|).
A cell is either blank or one of the agents.
The expression \verb|agent :> cell| means that an agent is a subtype of cell.

Let $\sqsubseteq$ be the reflexive transitive closure of \verb|:>|.
Let $\Sigma(f)$ be the type assigned to element $f$ by signature $\Sigma$.
Then $f(k_1, ..., k_n)$ is a well-formed term of type $t$ if:
\begin{itemize}
\item
$\Sigma(f) = (t_1, ..., t_n)$
\item
$\Sigma(k_i) \sqsubseteq t_i$ for all $i = 1..n$
\end{itemize}
Predicates are functions that return a \verb|bool| and constants are functions with no arguments.
For example, using the type signature above, \verb|true(at(3, 4, black))| is a well-formed term of type \verb|bool|, i.e. a well-formed ground atom.

\subsection{Automatically generating induction tasks for a GGP game}
\label{sec:generating-tasks}

Given a GGP game $\Gamma$ written in GDL, and a type signature $\Sigma$ for that game, our system automatically generates an IGGP induction task.
Before presenting the details, we summarise the general approach.
To generate the GGP dataset, we built a simple forward-chaining GDL interpreter.
We used the GDL interpreter to calculate the initial state, the currently valid moves, the transition function, and the reward.
When generating traces, we first calculate the actions that are currently available for each player.
Then we let each player choose actions uniform randomly.
We record the state trace $(s_1, ..., s_n)$, and extract a set of $(B_i, E^+_i, E^-_i)$  triples from each trace.
The target predicates we wish to learn are \verb|legal|, \verb|next|, \verb|goal|, and \verb|terminal|.
The $(B_i, E^+_i, E^-_i)$ triples for the predicates \verb|legal|, \verb|goal|, and \verb|terminal| are calculated from a single state, while the triples for \verb|next| are calculated from a pair of consecutive states $(s_i, s_{i+1})$.

We generated multiple traces for each game: 1000 episodes with a maximum of 100 time-steps.
However, we chose these numbers somewhat arbitrarily because there is a complex tradeoff on how much data to generate.
We want to generate enough data to capture the diversity of a game, so that a learner can (in theory) learn the correct game rules.
However, we do not want to generate too much data as to provide every game state, as this would mean that a learner would not need to learn anything, and could instead simply memorise game situations.
We also we do not want to generate too much data that it becomes expensive to compute or store.
It is, however, unclear where the boundary is between too little and too much data.
Whether such a boundary even exists itself is unclear because by imposing different biases, different learners may need more or less information on the same task.
In future work we would like to expand the dataset.
We then intend to repeat the experiments with different amounts of training data.

\begin{algorithm}
\DontPrintSemicolon
\SetKwInOut{Input}{input}
\SetKwInOut{Output}{output}
\Input{$\Gamma$, a GGP game written in the GDL language}
\Input{$\Sigma$, a type signature for $\Gamma$}
\Input{$\mathit{max}_\mathit{traces}$, the number of traces to generate}
\Input{$\mathit{max}_\mathit{time}$, the max number of time-steps in a trace}
\Output{a set of triples of the form $\{(B_i,E^+_i,E^-_i)\}^m_{i=1}$}
\BlankLine
\BlankLine
$\Lambda \leftarrow \{\}$ \;
\For{$i = 1 .. \mathit{max}_\mathit{traces}$}{
$s \leftarrow \mathit{initial}(\Gamma)$ \;
$t \leftarrow (s)$ \;
\For{$j = 2 .. \mathit{max}_\mathit{time}$}{
$s \leftarrow \mathit{next}(\Gamma, s)$ \;
$\mathit{append}(t, s)$ \;

\If{$\mathit{terminal}(\Gamma, s)$}
{
  $\mathit{break}$ \;
}

$\Lambda \leftarrow \Lambda \cup \mathit{extract}(t, \Sigma)$ \;

}
}
\Return{$\Lambda$}

\BlankLine
\BlankLine
\caption{Automatically generating induction tasks from GGP games}
\label{algo1}
\end{algorithm}

Our approach is presented in Algorithm \ref{algo1}.
This procedure generates a number of traces.
Each trace is a sequence of game states, and each game state is represented by a set of ground atoms.
We use the $\mathit{extract}$ function (described in Section \ref{sec:extract-function}) to produce a set of $(B_i,E^+_i,E^-_i)$ triples from a trace. We add this set of triples to $\Lambda$.
At the end, when we have finished all the traces, we return $\Lambda$, the set of triples.
The variable $s$ stores the current state (a set of ground atoms).
Initially, $s$ is set to the initial state: $\mathit{initial}(\Gamma)$ produces the initial state from the GDL description.
Then for each time-step, we calculate the next state via $\mathit{next}(\Gamma, s)$.
This function $\mathit{next}(\Gamma, s)$ involves three steps.
First, we calculate the available actions for each player.
Second, we let each player take a (uniform) random move.
Third, we use the transition function $T$ to calculate the next state from the current state $s$ and the actions of the players.
Once we have calculated the new state, we append it to the end of $t$.
Here, $t$ is a trace i.e. a sequence of states.
Then we check if the new state is terminal. If it is terminal, we finish the episode; otherwise, we continue for another time-step.
Once the episode is finished, we extract the set of $(B_i,E^+_i,E^-_i)$ triples from the sequence of states, and continue to the next trace.
Note that we need the type signature $\Sigma$ to extract the triples from the trace, but we do not need it to generate the trace itself.
For our experiments, we generated 1000 traces for each game, and ran for a maximum of 100 time-steps per game.

\subsubsection{The $\mathit{extract}$ function}
\label{sec:extract-function}

The $\mathit{extract}(t, \Sigma)$ function in Algorithm \ref{algo1} takes a trace $t = (s_1, ..., s_n)$ (a sequence of sets of ground atoms), and a type signature $\Sigma$ and produces a set of $(B_i,E^+_i,E^-_i)$ triples.
This set of triples represents a set of induction tasks for the distinguished predicates $\mathit{legal}$, $\mathit{goal}$, $\mathit{terminal}$, and $\mathit{next}$.
It is defined as:
\[
\mathit{extract}((s_1, ..., s_n), \Sigma) = \Lambda_1 \cup \Lambda_2 \cup \Lambda_3 \cup \Lambda_4
\]
where:
\begin{eqnarray*}
\Lambda_1 & = & \{\mathit{triple}_1(s_i, \mathit{legal}, \Sigma) \mid i = 1 .. n\} \\
\Lambda_2 & = & \{\mathit{triple}_1(s_i, \mathit{goal}, \Sigma) \mid i = 1 .. n\} \\
\Lambda_3 & = & \{\mathit{triple}_1(s_i, \mathit{terminal}, \Sigma) \mid i = 1 .. n\} \\
\Lambda_4 & = & \{\mathit{triple}_2(s_i, s_{i+1}, \Sigma) \mid i = 1 .. n-1\}
\end{eqnarray*}
Before we define the $\mathit{triple}_1$ and $\mathit{triple}_2$ functions, we introduce the relevant notation.
If $s$ is a set of ground atoms and $p$ is a predicate, let $s_p$ be the subset of atoms in $s$ that use the predicate $p$.
If $\Sigma$ is a type signature and $p$ is a predicate, then $\mathit{ground}(\Sigma, p)$ is the set of all ground atoms generated by $\Sigma$ that use predicate $p$.
Given this notation, we define $\mathit{triple}_1(s, p, \Sigma)$ = $(B, E^+, E^-)$ where:
\begin{eqnarray*}
& & B = s - s_p \\
& & E^+ = s_p \\
& & E^- = \mathit{ground}(\Sigma, p) - E^+
\end{eqnarray*}
To calculate the negative instances ${E}^-_i$, we use the closed-world assumption: all $p$-atoms not known to be true in $E^+$ are assumed to be false in $E^-$.
Given a type signature $\Sigma$, we generate the set $\mathit{ground}(\Sigma, p)$ of all possible ground atoms whose predicate is the distinguished predicate $p$.
For example, in a one player game, if $\mathit{ground}(\Sigma, \mathit{legal}) =  \{$\verb|legal(p1, up)|, \verb|legal(p1, down)|, \verb|legal(p1, left)|, and \verb|legal(p1, right)|$\}$, and $s_\mathit{legal}$ only contains \verb|legal(p1, up)| and \verb|legal(p1, down)|, then:
\begin{eqnarray*}
{E}^+_i & = & \{\verb|legal(p1, up)|, \verb|legal(p1, down)|\} \\
{E}^-_i & = & \mathit{ground}(\Sigma, \mathit{legal}) - {E}^+_i  = \{\verb|legal(p1, left)|, \verb|legal(p1, right)|\}
\end{eqnarray*}
We define $\mathit{triple}_2(s_i, s_{i+1}, \Sigma) = (B, E^+, E^-)$ where:
\begin{eqnarray*}
& & B = s_i \\
& & E^+ = s_{i+1} [\mathit{true} / \mathit{next}] \\
& & E^- = \mathit{ground}(\Sigma, \mathit{next}) - E^+
\end{eqnarray*}
When learning $\mathit{next}$, we use the facts at the earlier time-step $s_i$ as background facts, we use the facts at the later time-step $s_{i+1}$ as the positive facts $E^+$ to be learned (with the predicate $\mathit{true}$ replaced by $\mathit{next}$), and we use all the rest of the ground atoms involving $\mathit{next}$ as the negative facts $E^-$. Note, again, the use of the closed-world assumption: we assume all $\mathit{next}$ atoms not known to be in $E^+$ to be in $E^-$.

\section{Baselines and ILP systems}
\label{baselines}

We claim that IGGP is challenging for existing ILP approaches.
To support this claim we evaluate existing ILP systems on our IGGP dataset.
We compare the ILP systems against simple baselines.
We first describe the baselines and then each ILP system.

\subsection{Baselines}

Figure \ref{fig:baselines} shows the four baselines.
Each baseline is a Boolean function $f \colon 2^{\mathcal{B}} \times \mathcal{E} \to \{\top,\bot\}$, i.e. a function that takes background knowledge and an example and returns true ($\top$) or false $(\bot)$.
We describe these baselines in detail.

\begin{figure}[ht]
\normalsize
\centering
\begin{tabular}{l}
\hline
{\bf Baselines} \\ \hline
$True(B,a) = \top$ \\ \hline
$Inertia(B,a) = a[next/true] \in B$ \\ \hline
$Mean(B, a) = \{(B_i, E^+_i,E^-_i) \in \Delta \mid a \in E^+_i\} \geq \frac{|\Delta|}{2}$ \\ \hline
$KNN_k(B, a) = |\{ (B', E^{+'}, E^{-'}) \in \kappa_k(\Delta, B) \mid a \in E^{+'}\}| \geq \frac{\kappa_k(\Delta, B)}{2}$
\end{tabular}
\caption{Baselines where $\Delta = \{(B_i,E^+_i,E^-_i)\}^m_{i=1}$ represents training data. The syntax \tw{a[next/true]} means to replace the predicate symbol \tw{next} with \tw{true} in the atom $a$.}
\label{fig:baselines}
\end{figure}

\noindent
Our first two baselines ignore the training data:

\begin{itemize}
\item \textbf{True} deems that every atom is true:
$$True(B,a) = \top$$
\item \textbf{Inertia} is the same as \emph{True} for atoms with the target predicates \tw{goal}, \tw{legal}, and \tw{terminal}, but for the \tw{next} predicate an atom is true if and only if the corresponding \tw{true} atom is in $B$. For instance, the atom \tw{next(at(1,4,x))} is true if and only if \tw{true(at(1,4,x))} is in $B$:
$$Inertia(B,a) = a[next/true] \in B$$
The intuition behind this baseline is the empirical observation that in most of the games, most ground atoms retain their truth value from one time-step to the next, more often than not.
Of course, it is possible to design games in which most or all of the atoms change their truth value each time-step; but in typical games, such radical changes are unusual.
\end{itemize}


\noindent
Our next two baselines consider the training data $\Delta = \{(B_i,E^+_i,E^-_i)\}^m_{i=1}$:

\begin{itemize}
\item \textbf{Mean} deems that a testing atom $a$ is true if and only if $a$ is true more often than not in the positive training examples:
$$Mean(B, a) = |\{(B_i, E^+_i,E^-_i) \in \Delta \mid a \in E^+_i\}| \geq \frac{|\Delta|}{2}$$

\item \textbf{KNN$_k$} is based on clustering the data. In $KNN_k(B,a)$ we find the $k$ triples in $\Delta$, denoted as $\kappa_k(\Delta, B)$, whose backgrounds are most `similar' to the background $B$.
To assess the similarity of two sets $A$ and $B$ of ground atoms, we look at the size of the symmetric difference\footnote{For efficiency, we calculate this difference by converting the sets into bit vectors, applying xor, and counting the number of set bits.} between $A$ and $B$:
\[
d(A, B) = |A - B| + |B - A|
\]
It is straightforward to show that the $d$ function satisfies the conditions for a distance metric:
\begin{itemize}
\item
$d(A, B) \geq 0$
\item
$d(A, B) = d(B, A)$
\item
$d(A, B) = 0$ iff $A = B$
\item
$d(A, C) \leq d(A, B) + d(B, C)$
\end{itemize}
We set the closest $k$ triples $\kappa_k(\Delta, B)$ to be the $k$ triples $\{(B_i,E^+_i,E^-_i)\}^k_{i=1}$ with the smallest $d$ distance between $B_i$ and $B$.
Given the $k$ closest triples $\kappa_k(\Delta, B)$ the KNN baseline outputs $\top$ if $a$ appears in $E^{+'}$ in at least half of the closest $k$ triples. More formally:

$$KNN_k(B, a) = |\{ (B', E^{+'}, E^{-'}) \in \kappa_k(\Delta, B) \mid a \in E^{+'}\}| \geq \frac{k}{2}$$
\end{itemize}

\noindent
One potential limitation of the KNN approach is that, in contrast to the ILP approaches, the KNN approaches learn at the propositional level and are unable to learn general first-order rules. To illustrate this limitation, suppose we are trying to learn the target predicate $p/1$ given the background predicate $q/1$ and that the underlying target rule is $p(X) \leftarrow q(X)$. Suppose there are only two training triples of the form $(B,E^+,E-)$:

\begin{center}
  \begin{tabular}{l}
    $T_1 = (\{q(a)\}, \{ p(a) \}, \{ p(b), p(c) \})$\\
    $T_2 = (\{q(b)\}, \{ p(b) \}, \{ p(a), p(c) \})$
  \end{tabular}
\end{center}

\noindent
Given the test triple $(\{ q(c) \}, \{ p(c) \}, \{ p(a), p(b) \})$, a KNN approach will deem that $p(c)$ is false because it has not seen a positive instance of this particular ground atom and has no representational resources for generalising.

\subsection{ILP systems}

We evaluate four ILP systems on our dataset.
It is important to note that we are not trying to directly compare the ILP systems, or demonstrate that any particular ILP system is better than another.
We are instead trying to show that the IGGP problem is challenging for existing systems, and that it (and the dataset) will provide a challenging problem for evaluating future research.
Indeed, a direct comparison of ILP systems is often difficult \cite{crop:thesis}, largely because different systems excel at certain classes of problems.
For instance, directly comparing the Prolog-based Metagol against ASP-based systems, such as ILASP and HEXMIL \cite{hexmil} is difficult because Metagol is often used to learn recursive list manipulation programs, including string transformations and sorting algorithms \cite{crop:metaopt}.
By contrast, many ASP solvers disallow explicit lists, such as the popular Clingo system \cite{clingo:guide}, and thus a direct comparison is difficult.
Likewise, ASP-based systems can be used to learn non-deterministic specifications represented through choice rules and preferences modeled as weak constraints~\cite{law18generality}, which is not necessarily the case for Prolog-based systems.
In addition, because many of the systems have learning parameters, it is often possible to show that there exist some parameter settings for which system X can perform better than Algorithm Y on a particular dataset.
Therefore, the relative performances of the systems should largely be ignored.

We compare the ILP systems Aleph, ASPAL, Metagol, and ILASP.
We describe these systems in turn.



\subsubsection{Aleph}
\label{sec:aleph}

Aleph is an ILP system written in Prolog based on Progol \cite{mugg:progol}. Aleph uses the following procedure to induce a logic program hypothesis (paraphrased from the Aleph website\footnote{https://www.cs.ox.ac.uk/activities/programinduction/Aleph/}):

\begin{enumerate}
    \item Select an example to be generalised. If none exist, stop, otherwise proceed to the next step.
    \item Construct the most specific clause (also known as the bottom clause \cite{mugg:progol}) that entails the example selected and is within language restrictions provided.
    \item Search for a clause more general than the bottom clause. This step is done by searching for some subset of the literals in the bottom clause that has the `best' score.
    \item The clause with the best score is added to the current theory and all the examples made redundant are removed. Return to step 1.
\end{enumerate}

\noindent
To restrict the hypothesis space (mainly at step 2), Aleph uses both \emph{mode declarations} \cite{mugg:progol} and  \emph{determinations} to denote how and when a literal can appear in a clause.
In the mode language, \emph{modeh} are declarations for head literals and \emph{modeb} are declarations for body literals.
An example modeb declaration is \tw{modeb(2,mult(+int,+int,-int))}.
The first argument of a mode declaration is an integer denoting how often a literal may appear in a clause.
The second argument denotes that the literal mult/3 may appear in the body of a clause and specifies the type of its arguments.
The symbols $+$ and $-$ denote whether the arguments are input or output arguments respectively.
Determinations declare what predicates can be used to construct a hypothesis and are the form of \tw{determination(TargetName/Arity,BackgroundName/Arity)}.
The first argument is the name and arity of the target predicate.
The second argument is the name and arity of a predicate that can appear in the body of such clauses.
Typically there will be many determination declarations for a target predicate, corresponding to the predicates thought to be relevant in constructing hypotheses.
If no determinations are present Aleph does not construct any
clauses.

Aleph assumes that modes will be declared by the user. For the IGGP
tasks this is quite a burden because it requires that we create them for each
game, and also requires some knowledge of the target hypothesis we want to
learn. Fortunately, however, Aleph can extract mode declarations from
determinations, where determinations are straightforward to supply because we
can supply for each target predicate and each background predicate a
determination. Therefore, for each game, we allow Aleph to use all the
predicates available for that game as determinations and allow Aleph to induce
the necessary mode declarations.

There are many parameters in Aleph which greatly influence the output, such as
parameters that change the search strategy when generalising a bottom clause
(step 3) and parameters that change the structure of learnable programs (such
as limiting the number of literals in the bottom clause). We run Aleph using
the default parameters. Therefore, there will most likely exist some
parameter settings for which Aleph will perform better than we present.

We use Aleph 5 with YAP 6.2.2 \cite{yap}.

\subsubsection{ASPAL}

ASPAL~\cite{aspal} is a system for \emph{brave induction} under the answer set programming (ASP) \cite{lifschitz2008answer}
semantics.  Brave induction systems aim to find a hypothesis $H$ such that
there is at least one answer set of $B\cup H$ that covers the
examples\footnote{As the programs in this paper are guaranteed to be
\emph{stratified} -- recursion through negation is not allowed in this dataset
-- all programs have exactly one answer set and so the \emph{brave} and
\emph{cautious} settings for ILP under the answer set semantics coincide.}.

ASPAL works by transforming a brave induction task $T$ into a meta-level ASP
program $\mathcal{M}(T)$ such that the answer sets of $\mathcal{M}(T)$
correspond to the inductive solutions of $T$. The first step of
state-of-the-art ASP solvers, such as clingo~\cite{clingo}, is to compute the
\emph{grounding} of the program. Systems which follow this approach therefore
have scalability issues with respect to the size of the hypothesis space, as
every ground instance of every rule in the hypothesis space -- i.e.\ the ground
instances of every rule that has the potential to be learned -- is computed
when the ASP solver solves $\mathcal{M}(T)$.

Similarly to Aleph, ASPAL has several input parameters, which influence the size of the hypothesis space, such as the maximum number of body literals.
For most of these, we used the default value, but we increased the maximum number of body literals from 3 to 5 and the maximum number of rules in the hypothesis space from 3 to 15.
Our initial experiments showed that the maximum number of rules had very little effect on the feasibility of the ASPAL approach (as the size of the grounding of $\mathcal{M}(T)$ is unaffected by this change), whereas the maximum number of body literals can make a significant difference to the size of the grounding of $\mathcal{M}(T)$.
It is possible that there is a set of parameters for ASPAL that performs better than those we have chosen.

Predicate invention is supported in ASPAL by allowing new predicates (which do not occur in the rest of the task) to appear in the mode declarations.
This predicate invention is \emph{prescriptive} rather than automatic, as the schema of the new predicates (i.e.\ the arity, and argument types) must be specified in the mode declarations.
As how to guess the structure of predicates which should be invented is unclear for this problem setting, we did not allow ASPAL to use predicate invention on this dataset.
It should be noted that when programs are stratified, hypotheses containing predicate invention can always be translated into equivalent hypotheses with no predicate invention.
Of course, as such hypotheses may be significantly longer than the compact hypotheses which are possible through predicate invention, they may require more examples to be learned accurately by ASPAL.

Similarly, although ASPAL does enable learning recursive hypotheses, we did not permit recursion in these experiments.
Recursive hypotheses can also be translated into non-recursive hypotheses over finite domains.
Our initial experiments using ASPAL showed that in addition to increasing the size of the hypothesis space, allowing recursion also significantly increased the grounding of ASPAL's meta program, $\mathcal{M}(T)$.

\subsubsection{Metagol}
\label{sec:metagol}

Metagol \cite{mugg:metagold,crop:metafunc,metagol} is an ILP system based on a
Prolog meta-interpreter. The key difference between Metagol and a standard
Prolog meta-interpreter is that whereas a standard Prolog meta-interpreter
attempts to prove a goal by repeatedly fetching first-order clauses whose heads
unify with a given goal, Metagol additionally attempts to prove a goal by
fetching higher-order metarules (Figure \ref{fig:metarules}), supplied as
background knowledge, whose heads unify with the goal. The resulting
meta-substitutions are saved and can be reused in later proofs. Following the
proof of a set of goals, Metagol forms a logic program by projecting the
meta-substitutions onto their corresponding metarules.
Metagol is notable for its support for (non-prescriptive) predicate invention and learning recursive programs.

Metarules define the structure of learnable programs, which in turn defines the
hypothesis space. Deciding which metarules to use for a given task is an
unsolved problem~\cite{crop:thesis,crop:reduce}. To compute the benchmark, we set Metagol
to use the same metarules for all games and tasks. This set is composed of 9
derivationally irreducible metarules~\cite{crop:dreduce,crop:reduce}, a set of metarules to
allow for constants in a program, and a set of nullary metarules (to learn the
\tw{terminal} predicates). Full details on the metarules used can be found in
the code repository.

For each game, we allow Metagol to use all the predicates available for that
game. We also allow Metagol to support a primitive form of negation by
additionally using the negation of predicates. For instance, in
\emph{Firesheep} we allow Metagol to use the rule \tw{not\_does\_kill(A,B) :-
not(does\_kill(A,B))}. To allow Metagol to induce a program given all
$(B_i,E^+_i,E^-_i)$ triples, we prefix each atom with an extra argument to
denote which triple each atom belongs to. For instance, in the first
\emph{minimal even} triple, the atom \tw{does\_choose(player,1)} becomes
\tw{does\_choose(triple1,player,1)}, and in the second triple the same atom
becomes \tw{does\_choose(triple2,player,1)}. To account for this extra
argument, we also add extra argument to each literal in a metarule. For
instance, the \emph{ident} metarule becomes $P(I,A) \leftarrow Q(I,A)$ and the
\emph{chain} metarule becomes $P(I,A,B) \leftarrow Q(I,A,C), R(I,C,B)$.

We use Metagol 2.2.3 with YAP 6.2.2.

\begin{figure}[ht]
\centering
\begin{tabular}{l|l}
Name & Metarule \\ \hline
ident & $P(A,B) \leftarrow Q(A,B)$\\
curry & $P(A,B) \leftarrow Q(A,B,R)$\\
precon & $P(A,B) \leftarrow Q(A),R(A,B)$\\
chain & $P(A,B) \leftarrow Q(A,C),R(C,B)$\\
\end{tabular}
\caption{Example metarules. The letters $P$, $Q$, $R$ denote existentially quantified variables. The letters $A$, $B$, and $C$ denote universally quantified variables.}
\label{fig:metarules}
\end{figure}

\subsubsection{ILASP}
\label{sec:ilasp}

ILASP (Inductive Learning of Answer Set Programs)~\cite{law:ilasp,ILASP_system,law2015learning} is a collection of ILP systems, which are capable of learning ASP programs consisting of normal rules, choice rules, hard and weak constraints.
Unlike many other ILP approaches, ILASP guarantees the computation of an optimal inductive solution (where optimality is defined in terms of the length of a hypothesis).
Similarly to ASPAL, early ILASP systems, such as ILASP1~\cite{law:ilasp} and ILASP2~\cite{law2015learning}, work by representing an ILP task (i.e.\ every example and every rule in the hypothesis space) as a meta-level ASP program whose optimal answer sets correspond to the optimal inductive solutions of the task.
The ILASP systems each target learning unstratified ASP programs with normal rules, choice rules and both hard and weak constraints.
Therefore, the stratified normal logic programs which are targeted in this paper do not require the full generality of ILASP; in fact, on this dataset, the meta-level ASP programs used by both ILASP1 and ILASP2 are isomorphic to the meta-level program used by ASPAL.

ILASP2i~\cite{law2016iterative} addresses the scalability with respect to the number of examples by iteratively computing a subset of the examples, called relevant examples, and only representing the relevant examples in the ASP program.
In each iteration, ILASP2i uses ILASP2 to find a hypothesis $H$ that covers the set of relevant examples and then searches for a new relevant example which is not covered by $H$.
When no further relevant examples exist, the computed $H$ is guaranteed to be an optimal inductive solution of the full task.

Although ILASP2i makes significant improves on the scalability of ILASP1 and ILASP2 with respect to the examples, on tasks with large hypothesis spaces ILASP2i still suffers from the same grounding bottleneck as ASPAL, ILASP1 and ILASP2.
As the size of the hypothesis spaces are one of the major challenges of the dataset in this paper, ILASP2i would likely not perform significantly better than ASPAL.
To scale up the application of the ILASP framework to the GGP dataset, we used an extended version of ILASP2i, which computes, at each iteration, a \emph{relevant hypothesis space} using the type signature and the current set of relevant examples, and then uses ILASP2 to solve a learning task with the current relevant examples and relevant hypothesis space.
Through the rest of the paper, we refer to this extended ILASP algorithm as ILASP$^{*}$.
Specifically, rules that entail negative examples or do not cover at least one relevant positive example are omitted from the relevant hypothesis space.
Also, a rule is omitted if there is another rule which is shorter and covers the same (or more) relevant positive examples.
Similarly to ASPAL, ILASP$^{*}$ takes a parameter for the maximum number of literals in the body.
Our preliminary experiments showed that the method for computing the relevant hypothesis space performed best with this parameter set to 5, so this value was used for the experiments.

The construction of a relevant hypothesis space was made significantly easier by forbidding recursion and predicate invention in ILASP$^{*}$.
Although the standard ILASP algorithms do support recursion and (prescriptive) predicate invention, these two features mean that the usefulness of a rule in covering examples cannot be evaluated independently, and thus constructing the relevant hypothesis space is much more challenging.
In future work, we hope to generalise the method of relevant hypothesis space construction to relax these two constraints.

\section{Results}
\label{sec:results}

We now describe the results of running the baselines and ILP systems on our dataset.
All the experimental data is available at https://github.com/andrewcropper/mlj19-iggp.
When running the ILP systems, we allowed each system the same amount of time to learn each target predicate.
We allowed each system 30 minutes to learn each target predicate.

\subsection{Evaluation metrics}

We use two evaluation metrics: \emph{balanced accuracy} and \emph{perfectly solved}.

\subsubsection{Balanced accuracy}
\label{sec:ba}
In our dataset the majority of examples are negative.
To account for this class imbalance, we use \emph{balanced accuracy} \cite{balancedaccuracy} to evaluate the approaches.
Given background knowledge $B$, disjoint sets of positive $E^+$ and negative $E^-$ testing examples, and a logic program $H$, we define the number of positive examples as $p=|E^+|$, the number of negative examples as $n=|E^-|$, the number of true positives as $tp=|\{e \in E^+ | B \cup H \models e\}|$, the number of true negatives as $tn=|\{e \in E^- | B \cup H \not\models e\}|$, and the balanced accuracy $ba = (tp/p + tn/n)/2$.

\subsubsection{Perfectly solved}
\label{sec:ps}
We also consider a \emph{perfectly solved} metric, which is the number (or percentage) of tasks that an approach solves with 100\% accuracy.
The \emph{perfectly solved} metric is important in IGGP because we know that every game has at least one perfect solution: the GDL description from which the traces were generated is a perfectly accurate model of the deterministic MDP. Perfect accuracy is important because even slightly inaccurate models compound their errors as the game progresses.


\subsection{Results summary}

Figure \ref{fig:full-results} summarises the results and shows for each approach the balanced accuracy and percentage of perfectly solved tasks.
The full results are in the appendix.
As the results show, the ILP and KNN approaches perform better than simple baselines ($True$, $Inertia$, and $Mean$).
In terms of balanced accuracy, the KNN approaches often perform better than the ILP systems.
However, in terms of the important perfectly solved metric, the ILP methods easily outperform the baselines and the KNN approaches.
The most successful system ILASP$^{*}$ perfectly solves 40\% of the tasks.
It should be noted that 4\% of test cases have no positive instances in either the training set nor the test set, meaning that a perfect score can be achieved with the empty hypothesis.
Each of our ILP systems achieved a perfect score on these tasks.
Without these trivial cases, the score of each system on the perfectly solved metric would be even lower.

As Figure \ref{fig:target-ba} shows, in terms of balanced accuracies, the most difficult task is the \tw{terminal} predicate, although the margin of difference between the predicates is small.
As Figure \ref{fig:target-ps} shows, in terms of the important perfectly solved metric, the most difficult task is the \tw{next} predicate.
The mean number of perfectly solved tasks is a measly 3\%.
Even if we exclude the baselines and only consider the ILP systems then the mean is still only 10\%.
Figure \ref{sample-results} shows the balanced accuracies for the \tw{next} predicate on the alphabetically first ten games.
This predicate corresponds to the state transition function (Section \ref{sec:markov-games}).
The \tw{next} atoms are the most difficult to learn and there is only one out of the first ten games, \emph{Buttons and Lights}, for which any of the methods find a perfect solution.
The \tw{next} predicate is the most difficult to learn because it has the highest mean complexity in terms of the number of dependent predicates in the dependency graph (Section \ref{sec:gdl}) in the reference GDL game definitions.

\begin{figure}[ht]
\centering
\begin{tabular}{|l|c|c|c|c|c|c|c|c|c|}
\hline
\textbf{Metric} & \textbf{Baseline} & \textbf{Inertia} & \textbf{Mean} & \textbf{KNN$_1$} &\textbf{KNN$_5$} & \textbf{Aleph} & \textbf{ASPAL} & \textbf{Metagol}  & \textbf{ILASP$^{*}$} \\
\hline
BA (\%) & 48 & 56 & 64 & 80 & 80 & 66 & 55 & 69 & \textbf{86}\\
\hline
PS (\%) & 4 & 4 & 15 & 16 & 19 & 18 & 10 & 34 & \textbf{40}\\
\hline
\end{tabular}
\caption{
Results summary.
The baseline represents accepting everything.
The results show that all of the approaches struggle in terms of the perfectly solved metric (which represents how many tasks were solved with 100\% accuracy).
}
\label{fig:full-results}
\end{figure}


\begin{figure}[ht]
\centering
\begin{tabular}{|l|c|c|c|c|}
\hline
\textbf{Approach} & \textbf{\tw{goal}} & \textbf{\tw{legal}} & \textbf{\tw{next}} & \textbf{\tw{terminal}}\\
\hline
True & 47 & 56 & 47 & 42\\
Inertia & 47 & 56 & 80 & 42\\
Mean & 82 & 61 & 62 & 53\\
KNN$_1$ & \textbf{92} & 78 & 86 & 63\\
KNN$_5$ & \textbf{92} & 79 & 86 & 64\\
Aleph & 83 & 60 & 59 & 60\\
ASPAL & 52 & 59 & 50 & 59\\
Metagol & 74 & 66 & 60 & 77\\
ILASP$^{*}$ & \textbf{92} & \textbf{86} & \textbf{88} & \textbf{80}\\
\hline
Mean & 73 & 67 & 69 & 60\\
\hline
\end{tabular}
\caption{Balanced accuracy results for each target predicate.}
\label{fig:target-ba}
\end{figure}

\begin{figure}[ht]
\centering
\begin{tabular}{|l|c|c|c|c|}
\hline
\textbf{Approach} & \textbf{\tw{goal}} & \textbf{\tw{legal}} & \textbf{\tw{next}} & \textbf{\tw{terminal}}\\
\hline
True & 0 & 16 & 0 & 0\\
Inertia & 0 & 16 & 0 & 0\\
Mean & 32 & 16 & 0 & 12\\
KNN$_1$ & 34 & 16 & 0 & 12\\
KNN$_5$ & 34 & 22 & 0 & 18\\
Aleph & 32 & 18 & 4 & 16\\
ASPAL & 4 & 18 & 0 & 18\\
Metagol & \textbf{48} & 28 & 6 & \textbf{52}\\
ILASP$^{*}$ & 46 & \textbf{44} & \textbf{18} & \textbf{52}\\
\hline
Mean & 26 & 22 & 3 & 20\\
\hline
\end{tabular}
\caption{Perfectly solved percentage for each target predicate.}
\label{fig:target-ps}
\end{figure}

\begin{figure}[ht]
\centering
\begin{tabular}{|l|c|c|c|c|c|c|c|c|}
\hline
{\bf Game} & {\bf Inertia} & {\bf Mean} & {\bf KNN$_1$} & {\bf KNN$_5$} & {\bf Aleph} & {\bf ASPAL} & {\bf Metagol} & {\bf ILASP$^{*}$}\\ \hline
Alquerque             & {\bf 90}   & 73   & 87         & {\bf 90}   & 53   & 50   & 54   & 74 \\
Asylum                & 97         & 74   & {\bf 97}   & \bf{97}    & 69   & 50   & 51   & 84 \\
Battle of Numbers     & {\bf 88}   & 52   & 87         & 86         & 58   & 50   & 54   & 67 \\
Breakthrough          & {\bf 96}   & 70   & 95         & 96         & 52   & 50   & 51   & {\bf 97}\\
Buttons and Lights    & 54         & 50   & 82         & 81         & 58   & 50   & 50   & {\bf 100}\\
Centipede             & 67         & 57   & 88         & 85         & 57   & 50   & 50   & {\bf 92} \\
Checkers              & 91         & 66   & 90         & 90         & 55   & 50   & 55   & {\bf 95} \\
Coins                 & 79         & 50   & 88         & 81         & 63   & 50   & 60   & {\bf 93} \\
Connect 4 (Team)      & 93         & 50   & 92         & 92         & 50   & 50   & 50   & {\bf 96} \\
Don't Touch           & 89         & 76   & 86         & {\bf 90}   & 64   & 50   & 53   & 89 \\
\hline
\end{tabular}
\caption{Balanced accuracies for the \tw{next} target predicate for the alphabetically first ten games.}.
\label{sample-results}
\end{figure}

\noindent
In the following sections we analyse the results for each system and discuss the relative limitations of the respective systems on this dataset.

\subsubsection{KNN}

As Figure \ref{fig:full-results} shows, the KNN approaches perform well in terms of balanced accuracy but poorly in terms of perfectly solved. Note that KNN$_1$ occasionally scores higher than KNN$_5$, which is to be expected because sometimes looking at additional triples gives misleading information. As already mentioned, the KNN approaches learn at the propositional level. This limitation is evident when analysing the results which show that the KNN$_1$ and KNN$_5$ approaches only perform well when the target predicate can be learned by memorizing particular atoms. For some of the simpler games (e.g. \emph{Coins}), the KNN approach is often able to learn the \verb|goal| predicate because the reward can be extracted directly from the value of an internal state variable representing the score. Similarly, the KNN approach sometimes learns the \verb|legal| predicate when the set of legally valid actions is static and does not depend on the current state. But the KNN approach is not able to perfectly learn any of the \verb|next| rules for any of the games in our dataset. In addition, the KNN approaches are expensive to compute. To get these results it took 3 days on a 3.6 GHz machine.

\subsubsection{Aleph}

As Figure \ref{fig:full-results} shows, Aleph performs reasonably well, and outperforms most of the baselines in terms of the perfectly solved metric.
However, after inspecting the learned programs, we found that Aleph was rarely learning general rules for the games, and instead typically learned facts to explain the specific examples.
In other words, on this task, Aleph tends to learn overly specific programs.
There are several potential explanations for this limitation.
First, as we stated in Section \ref{sec:aleph}, we did not provide mode declarations to Aleph, and instead allowed Aleph to infer them from the determinations.
Second, we ran Aleph with the default parameters.
However, as stated in Section \ref{sec:aleph}, Aleph has many learning parameters which greatly influence the learning performance.
It is reasonable to assume that Aleph could perform even better with a different set of parameters.
Third, to learn a program Aleph must first construct the most specific clause (the bottom clause) that entails an example.
However, constructing the bottom clause requires exponential time in the depth of variables in the target theory \cite{mugg:progol}.
Therefore, learning large and complex clauses is intractable.

\subsubsection{ASPAL}

As Figure~\ref{fig:full-results} shows, ASPAL performs quite poorly on this dataset. It is outperformed by the \emph{mean} baseline, both in terms of the perfectly solved metric, and the average balanced accuracy.
ASPAL timed out on the majority of the test problems, which was caused by the size of the hypothesis space, and therefore the grounding of ASPAL's meta-level ASP program.
It is possible that by using different parameters to control the size of the hypothesis space, or using a different representation of the problem, with a smaller grounding, ASPAL could perform better.

The results of ASPAL are also interesting to explain the need to create a specialised version of the ILASP algorithm for this dataset.
On this constrained problem domain, where we are only aiming to learn stratified programs (which are guaranteed to have a single answer set), ILASP2 and ASPAL are almost identical in their approaches.
Both map the input ILP task into a meta-level ASP program, and use the Clingo ASP solver to find an optimal answer set, corresponding to an optimal inductive solution of the input task.
The specialised ILASP$^*$ algorithm presented in Section~\ref{sec:ilasp} can overcome this problem in some cases, by reducing the size of the hypothesis space being considered, and thus reducing the size of the grounding of the meta-level program.
In principle, this specialisation (along with ILASP2i's relevant example method) could be applied to ASPAL, to create ASPAL$^*$, which would likely have performed better.

\subsubsection{Metagol}
\label{sec:metagol-results}

Although Metagol outperforms the baselines in the perfectly correct metric (34\%), it is outperformed in terms of balanced accuracy.

One of the main limitations of Metagol in this dataset is that it will only return a program if that program covers all of the positive examples and none of the negative examples.
However, in some of the games, Metagol could learn a single simple rule that explains 99\% of the training examples (and perhaps 99\% of the testing examples) but may need an additional complex rule to cover the remaining 1\%.
If this extra rule is too complex to learn, then Metagol will not learn anything.
To explore this limitation we ran a modified version of Metagol that relaxes this constraint.
This modified version simply samples training examples, rather than learn from all the examples.
This stochastic version of Metagol improved balanced accuracy from 69\% to 76\%.
In future work we intend to develop more sophisticated versions of stochastic Metagol.

Metagol can generalise from few examples because of the strong inductive bias enforced by the metarules.
However, this strong bias is also a key reason why Metagol struggles to learn programs for many of the games.
Given insufficient metarules, Metagol cannot induce the target program.
For instance, given only monadic metarules, Metagol can only learn monadic programs.
Although there is work studying which metarules to use for monadic and dyadic logics \cite{crop:minmeta,crop:dreduce,crop:reduce}, there is no work on determining which metarules to use for higher-arity logic.
Therefore, when computing the benchmarks, Metagol could not learn some of the higher-arity target predicates, such as the \tw{next\_cell/4} predicate in \emph{Sudoku}.
Similarly Metagol could often not use higher-arity predicates, such as \tw{does\_move/5} and \tw{triplet/6} in \emph{Alquerque}.

Another issue with the metarules is in that, as described in Section \ref{sec:metagol}, we used the same set of metarules for all games.
This approach is inefficient because in almost all cases this approach meant that we were using irrelevant metarules, which added unnecessary search to the learning task.
We expect that a simple preprocessing step to remove unusable metarules would improve learning performance, although probably not by any considerable margin.

Another reason why Metagol struggles to solve certain games is because, as with most ILP systems, it struggles to learn large and complex programs.
For Metagol the bottleneck is in the size of the target program because the search space grows exponentially with the number of clauses in the target program \cite{crop:reduce}.
Although there is work in trying to mitigate this issue \cite{crop:metafunc}, developing approaches that can learn large and complex programs is a major challenge for MIL and ILP in general \cite{crop:thesis}.

\subsubsection{ILASP$^{*}$}

The system with the highest percentage of completely accurate models (see Figure \ref{fig:full-results}) is ILASP$^{*}$, with 40\% of the tasks completely solved.
In most of the cases where ILASP$^{*}$ terminated with a solution in the time limit of 30 minutes, a perfect solution was returned.
On the rare occasion that ILASP$^{*}$ terminated but learned an imperfect solution, it did cover the training examples, but performed imperfectly on the test set; for example, in the \texttt{terminal} training set for \texttt{Untwisty Corridor} there are no positive examples, meaning that ILASP$^{*}$ returns the empty hypothesis (which covers the set of negative examples); however, there is a positive instance of \texttt{terminal} in the test set, meaning that ILASP$^{*}$ (and all other approaches) score a balanced accuracy of 50 on this problem.

In some cases, the restriction on the number of body literals meant that the task had no solutions.
In these \emph{unsatisfiable} cases, the hypothesis in the last satisfiable iteration was returned by ILASP$^{*}$. In principle, the maximum number of body literals could have been iteratively increased until the task became satisfiable, but our initial experiments showed that this made little or no difference to the number of perfectly solved cases.
Some of the unsatisfiable cases may have been caused by the restriction forbidding predicate invention for ILASP$^{*}$ on this dataset -- although there will always by an equivalent hypothesis that does not contain predicate invention, the equivalent hypothesis may have rules with more than 5 body literals.

Similarly to the unsatisfiable cases, in the timeout cases, the hypothesis found in the ILASP$^{*}$'s final iteration was used to compute the accuracy.
Returning the hypothesis found in the last iteration explains ILASP$^{*}$'s much higher average balanced accuracy compared to Metagol, which either returns a perfect solution over the test set or no solution at all.

ILASP$^*$ is able to perfectly solve some tasks that are not perfectly solved by any of the baselines or other ILP systems. One example is the $\mathtt{next}$ learning task for \emph{Rock Paper Scissors}.
In this case, the raw hypothesis returned by ILASP$^*$ is shown in Figure \ref{fig:ilasp-rawh}, which is equivalent to the (more readable) hypothesis shown in Figure \ref{fig:ilasp-readh}.
Note that this hypothesis is slightly more complicated than necessary.
If ILASP$^*$ had been permitted to use $!=$ to check that two player variables did not represent the same player, it is possible that the last three rules would have been replaced with:

\begin{minted}{prolog}
next_score(Player1, Score) :-
  true_score(Player1, Score), does(Player1, Move1), does(Player2, Move2),
  not beats(Move1, Move2), Player1 != Player2.
\end{minted}

\noindent
It is possible to learn hypotheses with $!=$ (and other binary comparison operators) in ILASP, but this would have increased the size of the hypothesis space, so in these experiments, we only allowed ILASP$^*$ to construct hypothesis spaces using the language of the input task.
In future work, we may consider extending the relevant hypothesis space construction method to allow binary comparison operators.
The increase in the size of the hypothesis space may be outweighed by the fact that the final hypothesis can be shorter -- shorter hypotheses tend to need fewer iterations to learn.






\begin{figure}[ht]
\begin{minted}[frame=single]{prolog}
next_step(V0) :- succ(V2, V0), true_step(V2), int(V0), int(V2).

next_score(V0, V1) :-
  succ(V3, V1), beats(V8, V6), true_score(V0, V3), does(V5, V6), does(V0, V8),
  agent(V0), int(V1), int(V3), agent(V5), action(V6), action(V8).

next_score(V0, V1) :-
  true_score(V0, V1), does(V5, V7), does(V0, V7), V0 = p1, V5 = p2,
  agent(V0), int(V1), agent(V5), action(V7).

next_score(V0, V1) :-
  beats(V7, V8), true_score(V0, V1), does(V0, V8),
  does(V5, V7), agent(V0),
  int(V1), agent(V5), action(V7), action(V8).

next_score(V0, V1) :-
  true_score(V0, V1), does(V0, V8), does(V5, V8),
  V0 = p2, V5 = p1, agent(V0),
  int(V1), agent(V5), action(V8).
\end{minted}
\caption{The raw hypothesis returned by ILASP$^*$ for the \tw{next} learning task for \emph{Rock Paper Scissors}.}
\label{fig:ilasp-rawh}
\end{figure}

\begin{figure}[ht]
\begin{minted}[frame=single]{prolog}
next_step(NewStep) :- succ(CurrentStep, NewStep), true_step(CurrentStep).

next_score(Player1, NewScore) :-
  succ(Score, NewScore), true_score(Player1, Score), does(Player2, Move2),
  does(Player1, Move1), beats(Move1, Move2).

next_score(Player1, Score) :-
  true_score(Player1, Score), does(Player1, Move1), does(Player2, Move2),
  beats(Move2, Move1).

next_score(p1, Score) :-
  true_score(p1, Score), does(p2, Action), does(p1, Action).

next_score(p2, Score) :-
  true_score(p2, Score), does(p2, Action), does(p1, Action).
\end{minted}
\caption{A more readable version of the hypothesis returned by ILASP$^*$ for the \tw{next} learning task for \emph{Rock Paper Scissors}.}
\label{fig:ilasp-readh}
\end{figure}

\subsection{Discussion}

As Figure \ref{fig:full-results} shows, most of the IGGP tasks cannot be perfectly learned by existing ILP systems.
The best performing system (ILASP$^{*}$) solves only 40\% of the tasks perfectly.
Our results suggest that the IGGP problem poses many challenges to existing approaches.

As mentioned in Section \ref{sec:generating-tasks}, we are unsure whether the dataset contains sufficient training examples for each approach to perfectly solve all of the tasks.
Moreover, determining whether there is sufficient data is especially difficult because the different systems employ different biases.
However, in most cases the ILP systems simply timed out, rather than learning an incorrect solution.
The key issue is that the ILP systems we have considered do not scale to the large problems in the IGGP dataset.
In the previous section we discussed limitations of each system.
We now summarise the limitations to help explain what makes IGGP difficult for existing approaches.

\paragraph{Large programs}

As discussed in Section \ref{sec:rw}, many reference solutions for IGGP games are large, both in terms of the number of literals and the clauses in them.
For instance, the GGP reference solution for the \tw{goal} predicate for \emph{Connect Four} uses 14 clauses and a total of 72 literals.
However, learning large programs is a challenge for most ILP systems~\cite{crop:thesis} which typically struggle to learn programs with hundreds of clauses or literals.
Metagol, for instance, struggles to learn programs with more than 8 clauses.

\paragraph{Predicate invention}
The reference solution for \tw{goal} in \emph{Connect four} uses auxiliary predicates (\texttt{goal} is defined in terms of lines, which are defined in terms of columns, rows and diagonals).
These auxiliary predicates are not strictly required, as any stratified definition with auxiliary predicates can be translated into an equivalent program with no auxiliary predicates; however, such equivalent programs are often significantly longer.
If we unfold the reference solution to remove auxiliary predicates, the resulting equivalent unfolded program contains over 400 literals.
For ILP approaches that do not support the learning of programs containing auxiliary predicates (such as Progol, Aleph, and FOIL), it is infeasible to learn such a large program.
More modern ILP approaches support \emph{predicate invention}, enabling the learning of auxiliary predicates which are not in the language of the background knowledge or the examples; however, predicate invention is far from easy, and there are significant challenges associated with it, even for state of the art ILP systems.
ASPAL and ILASP support \emph{prescriptive} predicate invention, where the schema of the auxiliary predicates (i.e.\ the arity, and argument types) must be specified in the mode declarations~\cite{LawThesis}.
By contrast, Metagol supports \emph{automatic} predicate invention, where Metagol invents auxiliary predicates without the need for user-supplied arities or type information.
However, Metagol's approach can still often lead to inefficiencies in the search, especially when multiple new predicate symbols are introduced.





\section{Conclusion}

In this paper, we have expanded on the Inductive General Game Playing task proposed by Genesereth.
We claimed that learning the rules of the GGP games is difficult for existing ILP techniques.
To support this claim, we introduced a IGGP dataset based on 50 games from the GGP competition and we evaluated existing ILP systems on the dataset.
Our empirical results show that most of the games cannot be perfectly learned by existing systems.
The best performing system (ILASP$^{*}$) solves only 40\% of the tasks perfectly.
Our results suggest that the IGGP problem poses many challenges to existing approaches.
We think that the IGGP problem and dataset will provide an exciting challenge for future research, especially as we have introduced techniques to continually expand the dataset with new games.

\subsection{Limitations and future work}

\paragraph{Better ILP systems}

Our primary motivation for introducing this dataset is to encourage future research in ILP, especially on general ILP systems able to learn rules for a diverse set of tasks.
In fact, we have already demonstrated two advancements in this paper: (1) a stochastic version of Metagol (\ref{sec:metagol-results}), and (2) ILASP$^{*}$ (Section \ref{sec:ilasp}), which scales up ILASP2 for the GGP dataset.
In future work we intend to develop better ILP systems.

\paragraph{More games}
One of the main advantages of the IGGP problem is that the games are based on the GGP competition.
As mentioned in the introduction, the GGP competition produces new games each year.
These games are introduced independently from our dataset without any particular ILP system in mind.
Therefore, because of our second contribution, we can continually expand the IGGP dataset with these new games.
In future work we intend to automate this whole process and to ensure that all the data is publicly available.

\paragraph{More systems}
We have evaluated four ILP systems (Aleph, ASPAL, Metagol, and ILASP).
In future work we would like to evaluate more ILP systems.
We could also like to consider non-ILP systems (i.e. systems that may not necessarily learn explicit human-readable rules).

\paragraph{More evaluation metrics}
We have evaluated ILP systems according to two metrics: balanced accuracy and perfect solved.
However, there are other dimensions on which to evaluate the systems.
We have not, for instance, considered the learning times of the systems (although they all had the same maximum time to learn during the evaluation).
Nor have we considered the sample complexity of the approaches.
In future work it would be valuable to evaluate approaches when varying the number of game traces (i.e. observations) available, as to identify the most data-efficient approaches.

\paragraph{More challenges}
The main challenge in using existing systems on this dataset is the deliberate lack of game-specific language biases, meaning that for many games the hypothesis space that each system must consider is extremely large.
This reflects a major current issue in ILP, where systems are often given well crafted language biases to ensure feasibility; however, this is not the only current challenge in ILP.
For example, some ILP approaches target challenges such as learning from noisy data~\cite{HYPER_N,evans:dilp,law18noise}, probabilistic reasoning~\cite{de2007problog,de2010probabilistic,riguzzi2014history,bellodi2015structure,riguzzi2016s}, non-determinism expressed through unstratified negation~\cite{otero2001,law18generality}, and preference learning~\cite{law2015learning}.
Future versions of this dataset could be extended to contain these features.

\paragraph{Competitions}
SAT competitions have been held since 1992 with the aim of providing an objective evaluation of contemporary SAT solvers \cite{DBLP:journals/aim/JarvisaloBRS12}.
The competitions have significantly contributed to the progress of developing ever more efficient SAT techniques \cite{DBLP:journals/aim/JarvisaloBRS12}.
In addition, the competitions have motivated the SAT community to develop more robust, reliable, and general purposes SAT solvers (i.e implementations).
We believe that the ILP community stands to benefit from an equivalent competition, to focus and motivate research.
We hope that this new IGGP problem and dataset will become a central component in this new competition.

\newpage
\appendix

\section{Appendix: Full Results}
This appendix includes the full results for our dataset of 50 games. We use balanced accuracy as the evaluation metric (see Section \ref{sec:ba}).



\begin{longtable}{|l|l|l|l|l|l|l|l|l|l|l|}
\hline
  \textbf{Game} & \textbf{Predicate} & \textbf{True} & \textbf{Inertia} & \textbf{Mean} & \textbf{KNN(1)} & \textbf{KNN(5)} & \textbf{Aleph} & \textbf{ASPAL} & \textbf{Metagol} & \textbf{ILASP$^{*}$}
\\ \hline

Alquerque                        & \verb|goal|       & 50    & 50    & 50    & 97    & 95    & 100   & 50    & 100   & 100\\  \hline
Alquerque                        & \verb|legal|      & 50    & 50    & 52    & 62    & 63    & 50    & 50    & 50    & 63\\  \hline
Alquerque                        & \verb|next|       & 50    & 90    & 73    & 87    & 90    & 53    & 50    & 53    & 74\\  \hline
Alquerque                        & \verb|terminal|   & 50    & 50    & 50    & 51    & 50    & 49    & 50    & 100   & 100\\  \hline
Asylum                           & \verb|goal|       & 50    & 50    & 50    & 96    & 84    & 59    & 50    & 100   & 100\\  \hline
Asylum                           & \verb|legal|      & 50    & 50    & 52    & 73    & 69    & 50    & 50    & 50    & 62\\  \hline
Asylum                           & \verb|next|       & 50    & 97    & 74    & 97    & 97    & 68    & 50    & 51    & 84\\  \hline
Asylum                           & \verb|terminal|   & 0    & 0    & 100   & 99    & 100   & 98    & 100   & 100   & 100\\  \hline
Battle of Numbers                & \verb|goal|       & 50    & 50    & 74    & 98    & 97    & 50    & 50    & 50    & 73\\  \hline
Battle of Numbers                & \verb|legal|      & 50    & 50    & 56    & 68    & 67    & 50    & 50    & 50    & 78\\  \hline
Battle of Numbers                & \verb|next|       & 50    & 88    & 52    & 87    & 86    & 58    & 50    & 53    & 67\\  \hline
Battle of Numbers                & \verb|terminal|   & 50    & 50    & 50    & 53    & 50    & 48    & 50    & 50    & 39\\  \hline
Breakthrough                     & \verb|goal|       & 50    & 50    & 99    & 99    & 99    & 98    & 50    & 50    & 99\\  \hline
Breakthrough                     & \verb|legal|      & 50    & 50    & 51    & 73    & 70    & 50    & 50    & 50    & 73\\  \hline
Breakthrough                     & \verb|next|       & 50    & 96    & 70    & 95    & 96    & 51    & 50    & 51    & 97\\  \hline
Breakthrough                     & \verb|terminal|   & 50    & 50    & 50    & 50    & 50    & 49    & 50    & 50    & 52\\  \hline
Buttons and Lights               & \verb|goal|       & 50    & 50    & 83    & 83    & 83    & 100   & 50    & 50    & 90\\  \hline
Buttons and Lights               & \verb|legal|      & 100   & 100   & 100   & 100   & 100   & 100   & 100   & 100   & 100\\  \hline
Buttons and Lights               & \verb|next|       & 50    & 54    & 50    & 82    & 81    & 57    & 50    & 50    & 100\\  \hline
Buttons and Lights               & \verb|terminal|   & 50    & 50    & 50    & 100   & 100   & 75    & 50    & 100   & 100\\  \hline
Centipede                        & \verb|goal|       & 50    & 50    & 98    & 99    & 99    & 96    & 50    & 50    & 88\\  \hline
Centipede                        & \verb|legal|      & 50    & 50    & 86    & 73    & 95    & 78    & 50    & 50    & 91\\  \hline
Centipede                        & \verb|next|       & 50    & 67    & 56    & 88    & 85    & 56    & 50    & 50    & 92\\  \hline
Centipede                        & \verb|terminal|   & 50    & 50    & 50    & 89    & 82    & 52    & 50    & 50    & 75\\  \hline
Checkers                         & \verb|goal|       & 50    & 50    & 50    & 94    & 88    & 59    & 50    & 100   & 50\\  \hline
Checkers                         & \verb|legal|      & 50    & 50    & 54    & 64    & 62    & 50    & 50    & 50    & 75\\  \hline
Checkers                         & \verb|next|       & 50    & 91    & 66    & 90    & 90    & 55    & 50    & 55    & 95\\  \hline
Checkers                         & \verb|terminal|   & 50    & 50    & 50    & 50    & 60    & 48    & 50    & 50    & 74\\  \hline
Coins                            & \verb|goal|       & 50    & 50    & 100   & 100   & 100   & 93    & 50    & 100   & 100\\  \hline
Coins                            & \verb|legal|      & 50    & 50    & 50    & 66    & 50    & 49    & 50    & 50    & 56\\  \hline
Coins                            & \verb|next|       & 50    & 79    & 50    & 88    & 81    & 63    & 50    & 59    & 93\\  \hline
Coins                            & \verb|terminal|   & 50    & 50    & 50    & 83    & 92    & 68    & 50    & 50    & 95\\  \hline
Connect 4 (Team)                 & \verb|goal|       & 50    & 50    & 98    & 97    & 98    & 96    & 50    & 50    & 94\\  \hline
Connect 4 (Team)                 & \verb|legal|      & 50    & 50    & 62    & 50    & 66    & 55    & 50    & 50    & 92\\  \hline
Connect 4 (Team)                 & \verb|next|       & 50    & 93    & 50    & 92    & 92    & 50    & 50    & 50    & 96\\  \hline
Connect 4 (Team)                 & \verb|terminal|   & 50    & 50    & 50    & 49    & 50    & 49    & 50    & 50    & 58\\  \hline
Don't Touch                      & \verb|goal|       & 50    & 50    & 80    & 73    & 80    & 67    & 50    & 50    & 78\\  \hline
Don't Touch                      & \verb|legal|      & 50    & 50    & 50    & 68    & 89    & 49    & 50    & 73    & 100\\  \hline
Don't Touch                      & \verb|next|       & 50    & 89    & 76    & 86    & 90    & 64    & 50    & 53    & 89\\  \hline
Don't Touch                      & \verb|terminal|   & 50    & 50    & 50    & 47    & 49    & 51    & 50    & 50    & 100\\  \hline
Duikoshi                         & \verb|goal|       & 50    & 50    & 94    & 92    & 94    & 90    & 50    & 50    & 90\\  \hline
Duikoshi                         & \verb|legal|      & 50    & 50    & 51    & 73    & 79    & 49    & 50    & 50    & 70\\  \hline
Duikoshi                         & \verb|next|       & 50    & 93    & 59    & 92    & 92    & 52    & 50    & 52    & 94\\  \hline
Duikoshi                         & \verb|terminal|   & 50    & 50    & 50    & 49    & 50    & 52    & 50    & 50    & 57\\  \hline
Eight Puzzle                     & \verb|goal|       & 50    & 50    & 100   & 100   & 100   & 50    & 50    & 50    & 99\\  \hline
Eight Puzzle                     & \verb|legal|      & 50    & 50    & 50    & 92    & 82    & 51    & 50    & 50    & 100\\  \hline
Eight Puzzle                     & \verb|next|       & 50    & 84    & 52    & 89    & 88    & 49    & 50    & 55    & 86\\  \hline
Eight Puzzle                     & \verb|terminal|   & 50    & 50    & 50    & 50    & 50    & 51    & 50    & 100   & 100\\  \hline
Farming                          & \verb|goal|       & 50    & 50    & 98    & 100   & 99    & 100   & 50    & 100   & 100\\  \hline
Farming                          & \verb|legal|      & 50    & 50    & 52    & 66    & 82    & 49    & 50    & 50    & 100\\  \hline
Farming                          & \verb|next|       & 50    & 87    & 61    & 85    & 83    & 57    & 50    & 50    & 86\\  \hline
Farming                          & \verb|terminal|   & 50    & 50    & 50    & 79    & 85    & 49    & 50    & 100   & 100\\  \hline
Firesheep                        & \verb|goal|       & 50    & 50    & 63    & 97    & 97    & 100   & 50    & 100   & 100\\  \hline
Firesheep                        & \verb|legal|      & 50    & 50    & 78    & 86    & 84    & 49    & 50    & 50    & 97\\  \hline
Firesheep                        & \verb|next|       & 50    & 82    & 69    & 82    & 80    & 51    & 50    & 51    & 70\\  \hline
Firesheep                        & \verb|terminal|   & 50    & 50    & 50    & 90    & 92    & 48    & 50    & 50    & 38\\  \hline
Fizz-Buzz                        & \verb|goal|       & 50    & 50    & 100   & 100   & 100   & 100   & 50    & 50    & 100\\  \hline
Fizz-Buzz                        & \verb|legal|      & 50    & 50    & 88    & 90    & 89    & 86    & 50    & 100   & 100\\  \hline
Fizz-Buzz                        & \verb|next|       & 50    & 69    & 50    & 72    & 71    & 53    & 50    & 50    & 79\\  \hline
Fizz-Buzz                        & \verb|terminal|   & 50    & 50    & 50    & 50    & 50    & 48    & 50    & 100   & 100\\  \hline
Forager                          & \verb|goal|       & 50    & 50    & 50    & 97    & 100   & 55    & 50    & 100   & 100\\  \hline
Forager                          & \verb|legal|      & 100   & 100   & 100   & 100   & 100   & 100   & 100   & 100   & 100\\  \hline
Forager                          & \verb|next|       & 50    & 92    & 87    & 94    & 93    & 50    & 50    & 53    & 95\\  \hline
Forager                          & \verb|terminal|   & 50    & 50    & 50    & 50    & 47    & 46    & 50    & 100   & 61\\  \hline
Free For All                     & \verb|goal|       & 50    & 50    & 77    & 99    & 98    & 81    & 50    & 100   & 77\\  \hline
Free For All                     & \verb|legal|      & 50    & 50    & 52    & 76    & 72    & 50    & 50    & 50    & 96\\  \hline
Free For All                     & \verb|next|       & 50    & 86    & 65    & 86    & 84    & 59    & 50    & 58    & 63\\  \hline
Free For All                     & \verb|terminal|   & 50    & 50    & 50    & 61    & 54    & 52    & 50    & 100   & 100\\  \hline
Frogs and Toads                  & \verb|goal|       & 0    & 0    & 100   & 100   & 100   & 100   & 100   & 100   & 100\\  \hline
Frogs and Toads                  & \verb|legal|      & 50    & 50    & 50    & 95    & 87    & 50    & 50    & 50    & 74\\  \hline
Frogs and Toads                  & \verb|next|       & 50    & 95    & 93    & 97    & 97    & 51    & 50    & 51    & 85\\  \hline
Frogs and Toads                  & \verb|terminal|   & 0    & 0    & 100   & 100   & 100   & 100   & 100   & 100   & 100\\  \hline
GT Attrition                     & \verb|goal|       & 50    & 50    & 48    & 48    & 48    & 97    & 50    & 100   & 100\\  \hline
GT Attrition                     & \verb|legal|      & 50    & 50    & 0     & 50    & 0     & 100   & 50    & 100   & 100\\  \hline
GT Attrition                     & \verb|next|       & 38    & 60    & 54    & 64    & 62    & 57    & 50    & 78    & 86\\  \hline
GT Attrition                     & \verb|terminal|   & 50    & 50    & 50    & 50    & 50    & 100   & 50    & 100   & 100\\  \hline
GT Centipede                     & \verb|goal|       & 50    & 50    & 75    & 61    & 100   & 82    & 50    & 50    & 99\\  \hline
GT Centipede                     & \verb|legal|      & 50    & 50    & 50    & 75    & 100   & 0     & 50    & 100   & 100\\  \hline
GT Centipede                     & \verb|next|       & 43    & 61    & 53    & 69    & 64    & 59    & 50    & 71    & 100\\  \hline
GT Centipede                     & \verb|terminal|   & 0    & 0    & 0    & 50    & 0    & 100   & 100   & 100   & 100\\  \hline
GT Chicken                       & \verb|goal|       & 50    & 50    & 50    & 91    & 85    & 54    & 50    & 100   & 100\\  \hline
GT Chicken                       & \verb|legal|      & 50    & 50    & 50    & 75    & 86    & 49    & 50    & 100   & 100\\  \hline
GT Chicken                       & \verb|next|       & 50    & 59    & 50    & 79    & 71    & 50    & 50    & 67    & 68\\  \hline
GT Chicken                       & \verb|terminal|   & 50    & 50    & 50    & 57    & 70    & 46    & 50    & 100   & 100\\  \hline
GT Prisoner                      & \verb|goal|       & 50    & 50    & 50    & 83    & 78    & 56    & 50    & 100   & 100\\  \hline
GT Prisoner                      & \verb|legal|      & 50    & 50    & 50    & 93    & 100   & 49    & 50    & 100   & 100\\  \hline
GT Prisoner                      & \verb|next|       & 50    & 69    & 63    & 82    & 76    & 63    & 50    & 75    & 76\\  \hline
GT Prisoner                      & \verb|terminal|   & 50    & 50    & 50    & 80    & 94    & 46    & 50    & 100   & 100\\  \hline
GT Ultimatum                     & \verb|goal|       & 50    & 50    & 50    & 91    & 89    & 56    & 50    & 100   & 80\\  \hline
GT Ultimatum                     & \verb|legal|      & 50    & 50    & 61    & 95    & 100   & 69    & 50    & 69    & 100\\  \hline
GT Ultimatum                     & \verb|next|       & 45    & 61    & 68    & 68    & 71    & 57    & 50    & 61    & 84\\  \hline
GT Ultimatum                     & \verb|terminal|   & 50    & 50    & 50    & 75    & 78    & 52    & 50    & 100   & 100\\  \hline
Hex (Three)                      & \verb|goal|       & 50    & 50    & 100   & 100   & 100   & 99    & 50    & 50    & 99\\  \hline
Hex (Three)                      & \verb|legal|      & 50    & 50    & 53    & 47    & 56    & 50    & 50    & 50    & 52\\  \hline
Hex (Three)                      & \verb|next|       & 50    & 96    & 62    & 97    & 95    & 50    & 50    & 66    & 59\\  \hline
Hex (Three)                      & \verb|terminal|   & 50    & 50    & 50    & 50    & 50    & 49    & 50    & 50    & 45\\  \hline
Horseshoe                        & \verb|goal|       & 50    & 50    & 98    & 100   & 98    & 96    & 50    & 50    & 98\\  \hline
Horseshoe                        & \verb|legal|      & 50    & 50    & 55    & 94    & 77    & 57    & 50    & 50    & 100\\  \hline
Horseshoe                        & \verb|next|       & 50    & 64    & 50    & 87    & 83    & 69    & 50    & 65    & 90\\  \hline
Horseshoe                        & \verb|terminal|   & 50    & 50    & 50    & 78    & 67    & 55    & 50    & 50    & 77\\  \hline
Hunter                           & \verb|goal|       & 50    & 50    & 50    & 91    & 90    & 58    & 50    & 100   & 100\\  \hline
Hunter                           & \verb|legal|      & 50    & 50    & 50    & 90    & 83    & 53    & 50    & 50    & 100\\  \hline
Hunter                           & \verb|next|       & 50    & 88    & 77    & 88    & 90    & 69    & 50    & 52    & 87\\  \hline
Hunter                           & \verb|terminal|   & 50    & 50    & 50    & 62    & 59    & 46    & 50    & 100   & 100\\  \hline
Knights Tour                     & \verb|goal|       & 50    & 50    & 50    & 82    & 72    & 53    & 50    & 100   & 100\\  \hline
Knights Tour                     & \verb|legal|      & 50    & 50    & 50    & 73    & 63    & 51    & 50    & 50    & 77\\  \hline
Knights Tour                     & \verb|next|       & 50    & 83    & 64    & 87    & 84    & 63    & 50    & 50    & 94\\  \hline
Knights Tour                     & \verb|terminal|   & 50    & 50    & 50    & 45    & 63    & 52    & 50    & 50    & 54\\  \hline
Kono                             & \verb|goal|       & 50    & 50    & 50    & 97    & 95    & 100   & 50    & 100   & 100\\  \hline
Kono                             & \verb|legal|      & 50    & 50    & 53    & 61    & 65    & 50    & 50    & 50    & 82\\  \hline
Kono                             & \verb|next|       & 50    & 88    & 54    & 84    & 87    & 54    & 50    & 55    & 93\\  \hline
Kono                             & \verb|terminal|   & 50    & 50    & 50    & 52    & 53    & 51    & 50    & 97    & 97\\  \hline
Leafy                            & \verb|goal|       & 50    & 50    & 96    & 94    & 96    & 91    & 50    & 50    & 90\\  \hline
Leafy                            & \verb|legal|      & 50    & 50    & 50    & 56    & 56    & 50    & 50    & 50    & 100\\  \hline
Leafy                            & \verb|next|       & 50    & 97    & 90    & 96    & 97    & 49    & 50    & 100   & 92\\  \hline
Leafy                            & \verb|terminal|   & 50    & 50    & 50    & 51    & 50    & 49    & 50    & 50    & 47\\  \hline
Lightboard                       & \verb|goal|       & 50    & 50    & 100   & 100   & 100   & 100   & 50    & 50    & 98\\  \hline
Lightboard                       & \verb|legal|      & 100   & 100   & 100   & 100   & 100   & 100   & 100   & 100   & 100\\  \hline
Lightboard                       & \verb|next|       & 50    & 81    & 50    & 73    & 73    & 49    & 50    & 59    & 98\\  \hline
Lightboard                       & \verb|terminal|   & 50    & 50    & 50    & 45    & 50    & 48    & 50    & 100   & 100\\  \hline
Minimal Decay                    & \verb|goal|       & 0    & 0    & 100   & 100   & 100   & 100   & 100   & 100   & 100\\  \hline
Minimal Decay                    & \verb|legal|      & 100   & 100   & 100   & 100   & 100   & 100   & 100   & 100   & 100\\  \hline
Minimal Decay                    & \verb|next|       & 0    & 0    & 38    & 50    & 50    & 68    & 50    & 50    & 100\\  \hline
Minimal Decay                    & \verb|terminal|   & 50    & 50    & 100   & 100   & 100   & 100   & 100   & 100   & 100\\  \hline
Minimal Even                     & \verb|goal|       & 50    & 50    & 100   & 82    & 86    & 85    & 50    & 100   & 100\\  \hline
Minimal Even                     & \verb|legal|      & 100   & 100   & 100   & 100   & 100   & 100   & 100   & 100   & 100\\  \hline
Minimal Even                     & \verb|next|       & 50    & 89    & 50    & 84    & 87    & 100   & 50    & 100   & 100\\  \hline
Minimal Even                     & \verb|terminal|   & 50    & 50    & 50    & 69    & 75    & 100   & 50    & 100   & 100\\  \hline
Multiple Buttons and Lights      & \verb|goal|       & 50    & 50    & 100   & 100   & 100   & 100   & 50    & 100   & 100\\  \hline
Multiple Buttons and Lights      & \verb|legal|      & 100   & 100   & 100   & 100   & 100   & 100   & 100   & 100   & 100\\  \hline
Multiple Buttons and Lights      & \verb|next|       & 50    & 72    & 50    & 82    & 81    & 55    & 50    & 70    & 99\\  \hline
Multiple Buttons and Lights      & \verb|terminal|   & 50    & 50    & 50    & 98    & 100   & 48    & 50    & 100   & 100\\  \hline
Nine Board TicTacToe             & \verb|goal|       & 50    & 50    & 98    & 97    & 98    & 97    & 50    & 50    & 97\\  \hline
Nine Board TicTacToe             & \verb|legal|      & 6     & 6     & 54    & 67    & 60    & 50    & 50    & 50    & 85\\  \hline
Nine Board TicTacToe             & \verb|next|       & 5     & 54    & 55    & 98    & 97    & 52    & 50    & 97    & 94\\  \hline
Nine Board TicTacToe             & \verb|terminal|   & 50    & 50    & 50    & 49    & 50    & 49    & 50    & 50    & 52\\  \hline
Pentago                          & \verb|goal|       & 50    & 50    & 99    & 99    & 99    & 98    & 50    & 50    & 99\\  \hline
Pentago                          & \verb|legal|      & 50    & 50    & 52    & 67    & 65    & 50    & 50    & 56    & 85\\  \hline
Pentago                          & \verb|next|       & 50    & 91    & 50    & 87    & 84    & 52    & 50    & 53    & 94\\  \hline
Pentago                          & \verb|terminal|   & 50    & 50    & 50    & 50    & 50    & 49    & 50    & 50    & 64\\  \hline
Pilgrimage                       & \verb|goal|       & 50    & 50    & 100   & 100   & 100   & 99    & 50    & 50    & 70\\  \hline
Pilgrimage                       & \verb|legal|      & 50    & 50    & 50    & 65    & 65    & 49    & 50    & 50    & 70\\  \hline
Pilgrimage                       & \verb|next|       & 49    & 92    & 54    & 93    & 92    & 55    & 50    & 52    & 72\\  \hline
Pilgrimage                       & \verb|terminal|   & 0    & 0    & 0    & 0    & 0    & 100   & 100   & 100   & 100\\  \hline
Platform Jumpers                 & \verb|goal|       & 50    & 50    & 98    & 98    & 98    & 97    & 50    & 50    & 98\\  \hline
Platform Jumpers                 & \verb|legal|      & 50    & 50    & 52    & 66    & 62    & 50    & 50    & 55    & 83\\  \hline
Platform Jumpers                 & \verb|next|       & 50    & 98    & 76    & 99    & 99    & 56    & 50    & 50    & 73\\  \hline
Platform Jumpers                 & \verb|terminal|   & 50    & 50    & 50    & 50    & 74    & 48    & 50    & 50    & 30\\  \hline
Rainbow                          & \verb|goal|       & 50    & 50    & 99    & 99    & 99    & 97    & 50    & 50    & 95\\  \hline
Rainbow                          & \verb|legal|      & 50    & 50    & 50    & 81    & 86    & 50    & 50    & 100   & 47\\  \hline
Rainbow                          & \verb|next|       & 50    & 91    & 50    & 89    & 87    & 100   & 50    & 100   & 100\\  \hline
Rainbow                          & \verb|terminal|   & 50    & 50    & 50    & 46    & 50    & 57    & 50    & 50    & 80\\  \hline
Rock Paper Scissors              & \verb|goal|       & 50    & 50    & 50    & 75    & 79    & 100   & 50    & 100   & 100\\  \hline
Rock Paper Scissors              & \verb|legal|      & 100   & 100   & 100   & 100   & 100   & 100   & 100   & 100   & 100\\  \hline
Rock Paper Scissors              & \verb|next|       & 50    & 56    & 50    & 73    & 74    & 52    & 50    & 66    & 100\\  \hline
Rock Paper Scissors              & \verb|terminal|   & 50    & 50    & 50    & 100   & 100   & 0     & 50    & 100   & 100\\  \hline
Sheep and Wolf                   & \verb|goal|       & 50    & 50    & 100   & 100   & 100   & 50    & 50    & 100   & 56\\  \hline
Sheep and Wolf                   & \verb|legal|      & 41    & 41    & 55    & 65    & 66    & 50    & 50    & 50    & 54\\  \hline
Sheep and Wolf                   & \verb|next|       & 38    & 94    & 91    & 98    & 95    & 50    & 50    & 50    & 96\\  \hline
Sheep and Wolf                   & \verb|terminal|   & 0    & 0    & 0    & 0    & 0    & 98    & 100   & 100   & 100\\  \hline
Sokoban                          & \verb|goal|       & 50    & 50    & 50    & 50    & 50    & 49    & 50    & 50    & 72\\  \hline
Sokoban                          & \verb|legal|      & 50    & 50    & 50    & 72    & 75    & 53    & 50    & 50    & 71\\  \hline
Sokoban                          & \verb|next|       & 50    & 93    & 50    & 90    & 92    & 50    & 50    & 65    & 95\\  \hline
Sokoban                          & \verb|terminal|   & 50    & 50    & 50    & 50    & 50    & 49    & 50    & 50    & 50\\  \hline
Sudoku                           & \verb|goal|       & 50    & 50    & 100   & 100   & 100   & 100   & 50    & 100   & 100\\  \hline
Sudoku                           & \verb|legal|      & 50    & 50    & 53    & 98    & 97    & 50    & 50    & 50    & 55\\  \hline
Sudoku                           & \verb|next|       & 50    & 99    & 84    & 98    & 99    & 50    & 50    & 50    & 86\\  \hline
Sudoku                           & \verb|terminal|   & 50    & 50    & 50    & 49    & 50    & 48    & 50    & 50    & 48\\  \hline
Sukoshi                          & \verb|goal|       & 50    & 50    & 100   & 100   & 100   & 100   & 50    & 50    & 100\\  \hline
Sukoshi                          & \verb|legal|      & 50    & 50    & 50    & 87    & 91    & 60    & 50    & 50    & 91\\  \hline
Sukoshi                          & \verb|next|       & 50    & 93    & 65    & 90    & 93    & 69    & 50    & 50    & 93\\  \hline
Sukoshi                          & \verb|terminal|   & 50    & 50    & 50    & 44    & 49    & 43    & 50    & 50    & 43\\  \hline
Switches                         & \verb|goal|       & 0    & 0    & 100   & 100   & 100   & 100   & 50    & 100   & 100\\  \hline
Switches                         & \verb|legal|      & 50    & 50    & 50    & 84    & 85    & 52    & 100   & 50    & 89\\  \hline
Switches                         & \verb|next|       & 50    & 94    & 85    & 95    & 94    & 86    & 50    & 60    & 99\\  \hline
Switches                         & \verb|terminal|   & 0    & 0    & 100   & 100   & 100   & 100   & 100   & 100   & 100\\  \hline
TicTacToe                        & \verb|goal|       & 50    & 50    & 93    & 88    & 93    & 78    & 50    & 50    & 51\\  \hline
TicTacToe                        & \verb|legal|      & 50    & 50    & 53    & 72    & 91    & 48    & 50    & 72    & 100\\  \hline
TicTacToe                        & \verb|next|       & 50    & 85    & 51    & 83    & 91    & 54    & 50    & 55    & 89\\  \hline
TicTacToe                        & \verb|terminal|   & 50    & 50    & 50    & 64    & 57    & 45    & 50    & 50    & 71\\  \hline
Tiger vs Dogs                    & \verb|goal|       & 50    & 50    & 72    & 88    & 88    & 62    & 50    & 50    & 59\\  \hline
Tiger vs Dogs                    & \verb|legal|      & 50    & 50    & 50    & 57    & 64    & 50    & 50    & 50    & 79\\  \hline
Tiger vs Dogs                    & \verb|next|       & 50    & 91    & 72    & 89    & 92    & 51    & 50    & 51    & 54\\  \hline
Tiger vs Dogs                    & \verb|terminal|   & 0    & 0    & 100   & 100   & 100   & 100   & 100   & 100   & 100\\  \hline
Tron                             & \verb|goal|       & 50    & 50    & 50    & 75    & 71    & 29    & 50    & 50    & 91\\  \hline
Tron                             & \verb|legal|      & 50    & 50    & 50    & 80    & 84    & 54    & 50    & 50    & 85\\  \hline
Tron                             & \verb|next|       & 50    & 81    & 70    & 89    & 84    & 70    & 50    & 92    & 100\\  \hline
Tron                             & \verb|terminal|   & 50    & 50    & 50    & 70    & 77    & 56    & 50    & 50    & 100\\  \hline
TTCC4                            & \verb|goal|       & 50     & 50     & 100    & 100   & 100   & 100   & 50    & 50    & 100\\  \hline
TTCC4                            & \verb|legal|      & 23    & 23    & 52    & 75    & 66    & 50    & 50    & 50    & 74\\  \hline
TTCC4                            & \verb|next|       & 32    & 73    & 53    & 89    & 90    & 60    & 50    & 57    & 61\\  \hline
TTCC4                            & \verb|terminal|   & 0    & 0    & 100   & 98    & 100   & 97    & 100   & 100   & 71\\  \hline
Untwisty Corridor                & \verb|goal|       & 50    & 50    & 100   & 100   & 100   & 100   & 50    & 100   & 100\\  \hline
Untwisty Corridor                & \verb|legal|      & 100   & 100   & 100   & 100   & 100   & 100   & 100   & 100   & 100\\  \hline
Untwisty Corridor                & \verb|next|       & 50    & 76    & 80    & 92    & 91    & 75    & 50    & 61    & 100\\  \hline
Untwisty Corridor                & \verb|terminal|   & 50    & 50    & 50    & 50    & 50    & 50    & 50    & 50    & 50\\  \hline
Walkabout                        & \verb|goal|       & 50    & 50    & 95    & 95    & 95    & 92    & 50    & 50    & 93\\  \hline
Walkabout                        & \verb|legal|      & 50    & 50    & 50    & 71    & 82    & 51    & 50    & 50    & 100\\  \hline
Walkabout                        & \verb|next|       & 50    & 59    & 50    & 74    & 74    & 50    & 50    & 50    & 100\\  \hline
Walkabout                        & \verb|terminal|   & 50    & 50    & 50    & 50    & 50    & 50    & 50    & 100   & 100\\  \hline
\end{longtable}

\bibliographystyle{plain}
\bibliography{iggp}

\end{document}